\pdfoutput=1

\documentclass[journal]{IEEEtran}
\usepackage{amsfonts}
\usepackage{algorithm}
\usepackage{subfigure}

\usepackage{cite}

\ifCLASSINFOpdf
  \usepackage[pdftex]{graphicx}
\else
\fi
\usepackage{amsmath}
\usepackage{algorithmic}

\usepackage{array}

  \usepackage[caption=false,font=normalsize,labelfont=sf,textfont=sf]{subfig}
\usepackage{url}

\begin{document}
%
\title{Auto-weighted Multi-view Feature Selection with Graph Optimization}
%
%
%

\author{Qi Wang, \IEEEmembership{Senior Member, IEEE,} Xu Jiang, Mulin Chen and Xuelong Li, \IEEEmembership{Fellow, IEEE}
	\thanks{Q. Wang, X. Jiang, M. Chen and X. Li are with the School of Computer Science and School of Artificial Intelligence, Optics and Electronics (iOPEN), Northwestern Polytechnical University, Xi'an 710072, P.R. China (e-mail: crabwq@gmail.com; jx19961023@mail.nwpu.edu.cn; chenmulin@mail.nwpu.edu.cn; xuelong\_li@nwpu.edu.cn).
		}
	\thanks{\copyright 2021 IEEE. Personal use of this material is permitted. Permission from IEEE must be obtained for all other uses, in any current or future media, including reprinting/republishing this material for advertising or promotional purposes, creating new collective works, for resale or redistribution to servers or lists, or reuse of any copyrighted component of this work in other works.}
}

\markboth{IEEE TRANSACTIONS ON CYBERNETICS}%
{Shell \MakeLowercase{\textit{et al.}}: Bare Demo of IEEEtran.cls for IEEE Journals}
%



\maketitle

\begin{abstract}
In this paper, we focus on the unsupervised multi-view feature selection which tries to handle high dimensional data in the field of multi-view learning. Although some graph-based methods have achieved satisfactory performance, they ignore the underlying data structure across different views. Besides, their pre-defined laplacian graphs are sensitive to the noises in the original data space,  and fail to get the optimal neighbor assignment. To address the above problems, we propose a novel unsupervised multi-view feature selection model based on graph learning, and the contributions are threefold: (1) during the feature selection procedure, the consensus similarity graph shared by different views is learned. Therefore, the proposed model can reveal the data relationship from the feature subset. (2) a reasonable rank constraint is added to optimize the similarity matrix to obtain more accurate information; (3) an auto-weighted framework is presented to assign view weights adaptively, and an effective alternative iterative algorithm is proposed to optimize the problem. Experiments on various datasets demonstrate the superiority of the proposed method compared with the state-of-the-art methods.
\end{abstract}


%
\IEEEpeerreviewmaketitle

\section{Introduction}\label{sec:introduction}
\IEEEPARstart{T}{he} data representation of sample largely determines the performance of learning tasks. With the development of the information acquisition and storage technology, massive data is emerging. Multi-view data which contains heterogeneous features has shown greater superiority than the traditional single-view data, and it has received increasing attention in many scientific fields, e.g., coherent groups detection \cite{TPAMI}, \cite{MulinChenAAAI}, event detection \cite{SMDR}, image annotation \cite{MMFS}, \cite{reviewer31}, recognition and retrieval tasks \cite{ASVW5},  \cite{AMFS}, \cite{AUMFS}, \cite{ASVW16} and image clustering/classification tasks \cite{Caiguohao}, \cite{MCCDclass}, \cite{BBSMMclass}, \cite{HIDBclass}, \cite{MCNMFcluster}.
In some practical problems, an object can be described from many different aspects, which can be considered as its multiple view features. For example, in many applications, various visual features are extracted for image representation. The features from each view capture specific information of the image. Meanwhile, the underlying data structure is always shared by different views. Benefit from the diversity and consistency of different views, more useful and comprehensive information can be obtained, and it will further contribute to a better performance.

However, each view of the multi-view dataset is always with high dimensionality, which could lead to high computing cost, sparse data samples and so on. Moreover, the redundant features and noises contained in the original data space would also influence the final performance. As an efficient technique to alleviate these troubles, feature selection has attracted great attention in multi-view learning \cite{FSintro}. It aims to learn a compact and representative subset of original features, where the valuable information is preserved. Since the labels are always difficult to obtain, we are committed to the research of unsupervised multi-view feature selection in this paper. Generally, there are two different solutions for this task. 

The first way is to apply the traditional single-view feature selection approaches on multi-view data. Typically, the traditional single-view feature selection approaches include Structured Graph Optimization (SOGFS) \cite{SOGFS}, Joint Embedding Learning and Sparse Regression (JELSR) \cite{JELSR}, Robust Spectral Feature Selection (RSFS) \cite{RSFS}, Multi-Cluster Feature Selection \cite{MCFS}, Laplacian Score (LapScor) \cite{LAPSCOR}, and so on. These approaches rank features using different strategies, and are able to find the significant features in many applications. The traditional single-view approaches are specialized for single-view data. Therefore, it is not appropriate to apply them to multi-view data due to the lack of attention for potential relationship between different views. 

The second way is to use the methods designed specifically for multi-view data, where the underlying data structure across different views is valued. For example, Robust Multi-view Feature Selection (RMFS) \cite{RMFS} and Discriminatively Embedded K-Means (DEKM) \cite{DEKM} utilize the global structure to guide the multi-view subspace learning. Besides, owing to the powerful ability of graph learning \cite{reviewer32}, there are many graph-based approaches trying to preserve the local structure. Adaptive Unsupervised Multi-view Feature Selection (AUMFS) \cite{AUMFS} utilizes several vital information in a unified framework. Adaptive Multi-view Feature Selection (AMFS) \cite{AMFS} establishes a general trace ratio optimization model for human motion retrieval. AMFS and AUMFS characterize the local data structure by means of simply connecting pre-defined laplacian graph of each view. Similarly, Cluster Structure Preserving Unsupervised Feature Selection (CSP-UFS) \cite{CSP-UFS} construct the laplacian graph by adopting the same strategy as AMFS and AUMFS, but the difference is that it utilizes the discriminant analysis to preserve the cluster structure.

However, the original high-dimensional dataset always contains noise features and outliers. Importantly, the relationship between samples learned from the high-dimensional space can not capture its intrinsic characteristic \cite{adaptiveN}. Therefore, the pre-defined similarity graph is probably to be unreliable. Moreover, in multi-view data, different views explore the original data from different perspectives, and the real cluster structure exists in every view actually. Therefore, benefit from the consistency of different views, a unified similarity matrix can give more valuable and accurate information, and further contribute to a better performance. If the number of connected components equals to the number of clusters exactly, the similarity graph will have the ideal neighbor assignment. In clustering task, without further processing, a similarity graph with the ideal neighbor assignment can get the final clustering result directly. In the feature selection task, the guidance of a similarity graph is crucial to the process of feature selection. Accordingly, if the similarity graph has the ideal neighbor assignment, more accurate information will be obtained, and more valuable features will be selected. 
To address these issues, an unsupervised multi-view feature selection method named as Multi-view Feature Selection with Graph Learning (MFSGL) is proposed. We highlight the main contributions of the paper as follows:
\begin{enumerate}
	\item MFSGL learns an optimal similarity graph for all views, which indicates the cluster structure. A reasonable constraint is added to the similarity matrix, which make the number of connected components equals to the number of clusters exactly, and each connected component corresponds to one cluster.
	\item MFSGL carrys out the multi-view feature selection and similarity graph learning simultaneously. Therefore, it is able to learn the local data structure adaptively and get more valuable information.
	\item To balance the importance of different views, an efficient weight assignment strategy is proposed. The weight of each view can be determined more concisely and effectively.
\end{enumerate}
\par
The remainder of the paper is summarized here. Section \ref{sec:related work} reviews the related works on multi-view feature selection task. Section \ref{sec:Methodology} introduces the details of MFSGL. An effective optimization algorithm for this problem is also provided in Selection \ref{sec:Methodology}. In Section \ref{sec:EXPERIMENTS}, experiments on different datasets are conducted to validate the effectiveness of MFSGL, followed by the conclusion and future work in Section \ref{sec:Conclusion}.

\section{RELATED WORK}
\label{sec:related work}
Since MFSGL constructs the similarity graph to reveal the data cluster structure, in this section, we introduce several representative related works based on similarity graph constructing briefly
\subsection{AMFS}
Adaptive Multi-view Feature Selection (AMFS) is proposed for generating efficient motion data representation. Firstly, for preserving the local geometric structure, AMFS uses a local regression model and neighbor similarity weights to learn the laplacian graph of each view automatically. Secondly, these graphs are connected by non-negative view weights to explore the correlation of different views. Thirdly, to capture the global information, the final feature selection framework is formulated in a trace ratio form motivated by PCA. The final objective function is shown as:
\begin{equation}
\begin{split}
\min_{W,\alpha}\ &\frac{tr(W^TX\sum_{v=1}^{V}\alpha_v^rL^{(v)}X^TW)}{tr(W^TXH_XX^TW)}     \\
s.t.\ &\sum_{v=1}^V\,\alpha_v=1,\ \alpha_v \geq 0,\ W \in \{0,1\}^{D\times d},
\end{split}
\end{equation}
where $ \alpha_1,\alpha_2,\alpha_3,\cdots,\alpha_V $ are the non-negative view weights to combine all views, and $ r $ is used to prevent that only the best view is selected with its $ \alpha_v $ equals to 1. $ L^{(v)} $ denotes the pre-defined laplacian graph of $ v $-th view. $ W $ is the feature selection matrix that picks out most representative feature elements from original high-dimensional data. There is only one non-zero element in each row of $ W $, and the desired features will be identified according to the non-zero elements, so $ W^TX $ denotes the compact feature subset. $ D $ is the original feature number and is larger than $ d $. $ H $ is the centralized matrix.
\subsection{AUMFS}
Another method is Adaptive Unsupervised Multi-view Feature Selection (AUMFS) which is applied to several recognition tasks. It tries to simultaneously use the following key information: the local structure in data space, the similarity of different samples and the correlation of all views. Specifically, the authors attempt to obtain the pseudo cluster labels by a robust loss function based on regression model. It is necessary to consider the geometric structure hidden in the original data space, so the similarity graphs of each view are constructed.
What's more, to explore the underlying complemental information between different views, all views are connected using the non-negative view weights $ \lambda = [\lambda_1, \lambda_2, \cdots, \lambda_V]^T$. AUMFS is finally formulated in the following form:
\begin{equation}\label{equaumfs}
\begin{split}
&\min_{F,\lambda,W}\ tr(F^T\sum_{v=1}^{V}\lambda_v^r L^{(v)} F) + \alpha \lVert X^T W  - F \rVert_{2,1} + \beta \lVert W \rVert_{2,1}     \\
&s.t. \ F^TF=I_c, F \geq 0, \sum_{v=1}^{V}\lambda_v=1, \lambda_v \geq 0.
\end{split}
\end{equation}
The first term is used to obtain the data cluster labels and the last two terms make up the robust sparse regression model for feature selection. $ F $ denotes the predicted cluster label matrix. $ F^TF=I_c $ is an orthogonal constraint on $ F $. $ \lVert W \rVert_{2,1} $ is the feature selection matrix with the $ \ell_{2,1} $-norm regularization. $  X^T W $ can be regarded as the low-dimensional representation based on the most valued feature information. $ \lambda_v $ is the view weight vector. $ \alpha $ and $ \beta $ are used to balance the contributions of last two items. We can solve the problem by a simple efficient iterative method.
\subsection{MVFS}
Unsupervised Feature Selection for Multi-view Data (MVFS) \cite{MVFS} tries to explore the structural information existing in different views by utilizing pseudo cluster labels and minimizing the regression loss, which is similar to AUMFS. But it has two differences between MVFS and AUMFS. MVFS pre-defines the view weights which are fixed in the following iterations, and the another one is that MVFS uses the Frobenius norm in the regression term instead of the $ l_{2,1} $-norm.
\begin{equation}\label{equMVfs}
\begin{split}
\min_{W,Z}\ &\sum_{i=1}^{m} \lambda_i (Tr(Z^T L_i Z) + \alpha (\lVert X_i^T W_i  - Z \rVert_{F}^{2} + \beta \lVert W_i \rVert_{2,1} ) \\
s.t. \  &Z^T Z=I, Z \geq 0.
\end{split}
\end{equation}
Here, m is the view number. $Z$ denotes the pseudo label matrix. $L_i$ is the $i$-th laplacian matrix. $\lVert W_i \rVert_{2,1}$ controls the capacity and sparsity of the projection matrix $W_i$.  The parameter $\lambda_i$ denotes the $i$-th view weight. $ \alpha $ and $ \beta $ are two balanced parameters.
\subsection{CSP-UFS}
Cluster Structure Preserving Unsupervised Feature Selection (CSP-UFS) is proposed by Shi et al.. CSP-UFS tries to improve the ability of distinguishing different categories exactly after feature selection. The authors stress that the labels contain valuable information, so CSP-UFS uses spectral clustering to get the labels firstly. And then, the discriminant analysis is adopted to conduct the feature selection process while preserving the data cluster structure. Because the information of different views is complementary and reinforce each other, CSP-UFS imposes a non-negative weight on each view to connect all views, and the better view would get a larger weight. Therefore, CSP-UFS employs the spectral clustering and the discriminant analysis simultaneously for unsupervised feature selection.
\begin{equation}\label{equaumfs}
\begin{split}
\min_{W,F,\alpha} \ &\frac{Tr(W^{T}X(I-FF^{T})X^{T}W)}{Tr(W^{T}XX^{T}W)}\\ 
&+ \lambda_1 \sum_{v=1}^{V} \alpha_v^{r} Tr(F^{T}L^{(v)}F) + \lambda_2 \lVert W \rVert_{2,1}\\
s.t. \ &FF^{T}= I_c, F \geq 0, \sum_{v=1}^{V}\alpha_v=1, \alpha_v \geq 0.
\end{split}
\end{equation}
The first term is the feature selection procedure using discriminative analysis, and the last term ensures the sparsity of projection matrix $ W \in R^{d \times q} $. The pseudo label matrix $F$ can be learned in the second term. Similar to the above methods, $ L^{(v)} $ and $\alpha_v$ denote the $v$-th pre-defined laplacian matrix and view weight respectively. This problem can be solved by an alternating optimization algorithm.

\subsection{ASVW}
Obviously, all of the above methods pre-define the fixed laplacian graph of each view, and connect all views using a non-negative view weight vector. Different from them, Adaptive Similarity and View Weight (ASVW) \cite{ASVW} learns a similarity graph shared by different data views adaptively.
\begin{equation}
\begin{split}
\min \ &\mathbf{\mathcal{L}}(W_1, \cdots, W_V, \alpha, S) =   \\     
&\sum_{v=1}^{V} \sum_{i=1}^{n} \sum_{j=1}^{n} \alpha_v^{r_1} \lVert W_v^T x_i^{(v)}-W_v^T  x_j^{(v)} \rVert^2 (s_{ij})^{r_2}\\
&+ \lambda\sum_{v=1}^{V} \lVert W_v \rVert_{2,p}^p    \\
s.t. \  &W_v^TW_v=I, \sum_{v=1}^{V}\alpha_v=1, \alpha_v \geq 0, \sum_{j=1}^{n} s_{ij}=1,\\
&s_{ij} \geq 0, \lVert s_i \rVert_0 = k.
\end{split}
\end{equation}	
Here, $ S $ and $ \alpha_v $ denote the similarity matrix and $ v $-th view weight respectively. $ r_1 $ and $ r_2 $ are two parameters to avoid the trivial solution. The $ \ell_{2,p} $-norm is used to ensure the sparsity of projection matrix $ W_v $. In the first term, the common similarity matrix is learned adaptively. However, the similarity matrix learned by ASVW fails to have an ideal neighbor assignment, and the clustering result can not be obtained directly from it.

\section{Methodology}
\label{sec:Methodology}
In this section, we first formulate the objective function of MFSGL, and then an efficient alternative iterative algorithm is introduced to solve it.
\par
For better introducing the proposed method, all the notations are summarized in Table \ref{tab:notations}.
%
%

\begin{table}[htbp] 
	\centering
	\caption{\label{tab:test}Descriptions of all notations.} 
	\setlength{\tabcolsep}{7mm}{
		\begin{tabular}{ccccc}
			\hline	
			notations & descriptions\\
			\hline
			$n$&the number of samples \\
			$c$&the number of clusters\\
			$V$&the number of views\\
			$s$&the number of selected features\\
			$d_v$&the dimension in the $v$-th view\\
			$m_v$&the projection dimension in the $v$-th view\\
			$\alpha^{(v)}$&the $ v $-th view weight\\
			$x_i^{(v)} \in \mathbb{R}^{d_v}$&the $i$-th sample in the $v$-th view\\
			$X^{(v)} \in \mathbb{R}^{d_v \times n}$&the $v$-th view of the multi-view data\\
			$W_v \in \mathbb{R}^{d_v \times m_v} $&the projection matrix of the $v$-th view\\
			$S$&the similarity matrix\\
			$F$&the cluster indicator matrix\\
			$ Tr(\cdot) $&the trace of a matrix\\
			$\lVert \cdot \rVert _F$ & Frobenius norm  \\
			$\lVert \cdot\rVert_{2}$ & $\ell_{2}$-norm of a vector  \\
			$\lVert \cdot\rVert_{2,1}$ & $\ell_{2,1}$-norm  \\
			$ \gamma, \lambda, \mu $ &the regularization parameters\\
			\hline
	\end{tabular} }
	\label{tab:notations}
\end{table}


\subsection{Multi-View Feature Selection with Graph Learning}
\label{subsec: Multi-View Feature Selection with Graph Learning}
The local structure is pretty conspicuous for its information discovery capability, and it is believed better than the global structure. Therefore, lots of unsupervised feature selection methods explore the local structure information and preserve it after projection. The traditional methods always learn the similarity graph of each view in advance. However, as mentioned above, the pre-defined similarity graph may be unreliable. In fact, to efficiently capture the structure of data space, we propose to learn the graph and the feature subset simultaneously. In the data space, the closer the two data points are, the larger similarity between them should be. In this paper, the square of Euclidean distance is used as the measurement of the similarity between samples. To evaluate the effectiveness of different views, a reasonable view weight assignment method is also required. 
\par Based on above discussion, we firstly introduce the following  unsupervised multi-view feature selection framework:
\begin{equation}\label{1}
\begin{split}
\min \limits_{W_v, S}\  &\sum_{v=1}^{V} (({\sum_{i,j}\lVert W_{v}^T x_i^{(v)}-W_{v}^T x_j^{(v)} \rVert_2^2 s_{ij} })^{ \frac{p}{2}} + \gamma \lVert W_v \rVert_{2,1})\\
s.t. \ &\forall i, s_i^T \mathbf{1} = 1, 0 \leq s_{ij} \leq 1, W_v^T W_v=I.
\end{split}
\end{equation}
Here, for $ v$-th view, the projection matrix is denoted as $ W_{v} \in \mathbb{R}^{d_v \times m_v} $, and it projects the original $d_v$-dimensional data space into the latent $m_v$-dimensional subspace, where $m_v$ less than $d_v$ definitely. Moreover, in experiments, $m_v$ always needs to be tuned to get the best result \cite{SOGFS}. $ \gamma $ is the regularization parameter. The $ \ell_{2,1} $-norm regularization on $ W $ makes it row sparse to select more valuable features \cite{Wsparse}, \cite{Wsparse2}. The constraint $  W_v^T W_v=I $ makes the feature space distinctive after reduction \cite{Wdistinctive}. For similarity matrix $ S \in \mathbb{R}^{ n \times n } $, the element $ s_{ij} $ denotes the similarity between \textit{i}-th sample and \textit{j}-th sample, and the transpose of the $i$-th row is denoted as $ s_{i} $. We denote the column vector whose all elements are 1 as $\mathbf{1}$. $\lVert W_{v}^T x_i^{(v)}-W_{v}^T x_j^{(v)} \rVert_2^2$ denotes the distance between two samples in \textit{v}-th view after reduction. With different parameter $ p $ ($ 0 < p \leq 2 $), different exponential functions are obtained in Eq. (\ref{1}). Thanks to the exponential function, the weight of each view can be assigned automatically which will be presented in the following part.
\par
Denote the laplacian matrix as $ L_S = D- \frac{S^T+S}{2}$, and the elements of diagonal matrix $ D $ are set as $ \sum_j \frac{(s_{ij}+s_{ji})}{2} $. It can be proved that Eq. (\ref{1}) is equivalent to
\begin{equation}\label{i1}
\begin{split}
\min \limits_{W_v, S}\  &\sum_{v=1}^{V} ( \alpha_v Tr(W_{v}^T (X^{(v)})^T L_S X^{(v)} W_{v}) + \gamma \lVert W_v \rVert_{2,1})\\
s.t. \ &\forall i, s_i^T \mathbf{1} = 1, 0 \leq s_{ij} \leq 1, W_v^T W_v=I,
\end{split}
\end{equation}
where
\begin{equation}\label{i2}
\begin{split}
\alpha_v =\frac{p}{2 {Tr(W_{v}^T (X^{(v)})^T L_S X^{(v)} W_{v}) }^{\frac{2-p}{2}}}.
\end{split}
\end{equation}

\par
\textit{Proof}: The Eq. (\ref{1}) can be rewritten as
\begin{equation}\label{1a}
\begin{split}
\min \limits_{W_v, S}\  &\sum_{v=1}^{V}  ({Tr(W_{v}^T (X^{(v)})^T L_S X^{(v)} W_{v}) }^{\frac{p}{2}} + \gamma \lVert W_v \rVert_{2,1})\\
s.t. \ &\forall i, s_i^T \mathbf{1} = 1, 0 \leq s_{ij} \leq 1, W_v^T W_v=I.
\end{split}
\end{equation} 
We denote the Lagrangian multiplier as $\Lambda$ and the Lagrangian function of problem (\ref{1a}) should be
\begin{equation}\label{3}
\begin{split}
&\sum_{v=1}^{V} ({Tr(W_{v}^T (X^{(v)})^T L_S X^{(v)} W_{v}) }^{\frac{p}{2}} + \gamma \lVert W_v \rVert_{2,1}\\
&+\mathcal{G}(\Lambda,W_v,S)).
\end{split}
\end{equation}
Then, let the derivative of Eq. (\ref{3}) w.r.t $ W_v $ be zero, and we get
\begin{equation}\label{4}
\begin{split}
&\sum_{v=1}^{V} (\alpha_v \frac{\partial Tr(W_{v}^T (X^{(v)})^T L_S X^{(v)} W_{v})}{\partial W_v}+ \gamma\frac {\partial \lVert W_v \rVert_{2,1}}{\partial W_v}+\\&\frac {\partial\mathcal{G}(\Lambda,W_v,S)}{\partial W_v})=0
\end{split}
\end{equation}
and Eq. (\ref{i2}).
\par
If $ \alpha_v $ is set to be stationary, solving Eq. (\ref{4}) is equivalent to solving Eq. (\ref{i1}). Therefore, the solution of Eq. (\ref{1}) is transformed into the alternative iterative solution of Eq. (\ref{i1}) and Eq. (\ref{i2}).
If the feature set of $ v $-th view is more compact and efficient, $\sum_{i,j} \lVert W_{v}^T x_i^{(v)}-W_{v}^Tx_j^{(v)} \rVert_2^2 s_{ij}$ ought to be smaller, which brings a larger $ v $-th view weight $ \alpha_v $ based on Eq. (\ref{i2}). Accordingly, a worse view will be assigned a smaller weight. Therefore, the introduced framework is able to optimize the weights adaptively.
\par 
If the number of connected components in a similarity matrix equals to $ c $, the cluster structure can be discovered clearly, and it is conducive to the follow-up treatment obviously in clustering tasks. However, the similarity matrix learned by problem (\ref{1}) fails to have this property \cite{16}. Fortunately, if the rank of $ L_S$ is equivalent to $n-c$, this problem will be solved \cite{rankconstraint,CLR}. $ L_S$ is positive semi-definite, that is to say, $\sigma_i(L_S) \geq 0 $ where $\sigma_i(L_S)$ denotes the $ i $-th smallest eigenvalue of $ L_S $.
It is able to be proved that $ rank(L_S)=n-c $ brings $\sum_{i=1}^{c} \sigma_i(L_S) =0$.
And then, based on Ky Fan’s Theorem \cite{ky}, the following equation is obtained: 
\begin{equation}\label{5}
\begin{split}
\sum_{i=1}^{c} \sigma_i(L_S) = \min_{F^TF=I, F \in \mathbb{R}^{n \times c}}Tr(F^T L_S F),
\end{split}
\end{equation}
where $F$ denotes the cluster indicator matrix.
Moreover, to avoid the trivial solution, the constraint $\mu \sum_{i,j}s_{ij}^2$ should be added where $\mu$ is a regularization parameter \cite{SEAN}. Otherwise, the optimal solution should be that the similarity between the closest two samples is 1, and the others are 0.
\par 
By combining all of the above constraints, we finally summarize the Unsupervised Multi-view Feature Selection with Graph Learning (MFSGL) framework as
\begin{equation}\label{final}
\begin{split}
\min \limits_{W_v, F, S}\  &\sum_{v=1}^{V} ({Tr(W_{v}^T (X^{(v)})^T L_S X^{(v)} W_{v}) }^{\frac{p}{2}} + \gamma \lVert W_v \rVert_{2,1})\\&+ \mu \sum_{i,j}s_{ij}^2+2 \lambda Tr(F^T L_S F)\\
s.t. \ &\forall i, s_i^T \mathbf{1} = 1, 0 \leq s_{ij} \leq 1, F^TF=I, F \in \mathbb{R}^{n \times c}\\& W_v^T W_v=I.
\end{split}
\end{equation}
Here, the regularization parameter $\lambda$ is used to guarantee the existence of ideal neighbor assignment. The value of $\lambda$ can be determined adaptively during iteration. If there are more than $c$ connected components in $S$, $\lambda$ should be decreased. Otherwise, $\lambda$ should be increased. In summary, we integrate the feature selection and similarity graph learning, and the similarity matrix is constrained to have exactly $c$ connected components by the last term. Therefore, the proposed method MFSGL can learn an optimal similarity graph and a reliable projection matrix.
\par
After deriving the optimal solution, $\lVert w_{vi} \rVert_{2}$ is adopted to evaluate the importance of each feature of all views, where $w_{vi}$ denotes the $ i $-th row of $ W_v $. 
And the \textit{s} features we want to select are the top \textit{s} features based on $\lVert w_{vi} \rVert_{2}$.

\subsection{Optimization Algorithm}
From section \ref{subsec: Multi-View Feature Selection with Graph Learning}, it is easy to know that the solution of problem (\ref{final}) can be transformed into the alternative iterative solution of the following problem and Eq. (\ref{i2}).
\begin{equation}\label{2}
\begin{split}
\min \limits_{W_v, F, S}\  &\sum_{v=1}^{V} ( \alpha_v Tr(W_{v}^T (X^{(v)})^T L_S X^{(v)} W_{v}) + \gamma \lVert W_v \rVert_{2,1})\\&+ \mu \sum_{i,j}s_{ij}^2+2 \lambda Tr(F^T L_S F)\\
s.t. \ &\forall i, s_i^T \mathbf{1} = 1, 0 \leq s_{ij} \leq 1, F^TF=I, F \in \mathbb{R}^{n \times c}\\& W_v^T W_v=I.
\end{split}
\end{equation}
With fixed $ \alpha_v $, the solution of problem (\ref{2}) is the same as the solution of problem (\ref{final}), and we can get $ F$, $ W_v $ and $ S $ by solving problem (\ref{2}). Then, according to the newly obtained $ W_v $ and $ S $, $ \alpha_v  $ is able to be updated by Eq. (\ref{i2}).
\par 
Here, an efficient alternative iterative algorithm is given to get the solution of problem (\ref{2}).
\subsubsection{Update $ W_v $ with Fixed $ \alpha_v $ , F and S}
If $\alpha_v$, $F$ and $S$ are fixed, we can rewrite the problem (\ref{2}) as the following problem for each view:
\begin{equation}\label{W1}
\begin{split}
\min_{W_v^TW_v=1} \    \alpha_vTr(W_{v}^T (X^{(v)})^T L_S X^{(v)} W_{v}) + \gamma \lVert W_v \rVert_{2,1},
\end{split}
\end{equation}
where $ \lVert W_v \rVert_{2,1} = \sum_i\lVert w_{vi} \rVert_{2} $. Apparently, $  w_{vi} $ is theoretically possible to be zero, which will cause Eq. (\ref{W1}) non-differentiable. So we set $ \varepsilon $ as a very small constant and replace $ \lVert w_{vi} \rVert_{2} $ with $\sqrt{{w_{vi}}^T w_{vi}+\varepsilon}$. Then, problem (\ref{W1}) is converted into

\begin{equation}\label{W2}
\begin{split}
\min_{W_v^T W_v=1} \  &\alpha_v Tr(W_{v}^T (X^{(v)})^T L_S X^{(v)} W_{v}) \\ &+ \gamma \sum_{i} \sqrt{w_{vi}^Tw_{vi}+\varepsilon}.
\end{split}
\end{equation}
If $\varepsilon  $ approaches zero infinitely, the problem (\ref{W1}) is the same as problem (\ref{W2}). 
\par 
We denote the Lagrangian multiplier as $  \Lambda $ and write the Lagrangian function of problem (\ref{W2}) as
\begin{equation}\label{W3}
\begin{split}
&\mathbf{\mathcal{L}}(W_v,\Lambda) =       
\alpha_v Tr(W_{v}^T (X^{(v)})^T L_S X^{(v)} W_{v})\\&+ \gamma \sum_i \sqrt{w_{vi}^Tw_{vi}+\varepsilon}
+Tr(\Lambda(W_v^TW_v-1)),
\end{split}
\end{equation}	
Then, take the derivative of Eq. (\ref{W3}) on $ W_v $ and let it be zero:
\begin{equation}\label{W4}
\begin{split}
\frac{\partial\mathbf{\mathcal{L}}(W_v,\Lambda)}{\partial W_v} =        
\alpha_v  (X^{(v)})^T L_S X^{(v)} W_{v}+ \gamma G W_v
+W_v \Lambda=0,
\end{split}
\end{equation}	
where $ G \in \mathbb{R}^{d_v \times d_v}$ is a diagonal matrix whose diagonal entries are defined as 
\begin{equation}\label{G}
\begin{split}
G_{ii}= \frac{1}{2\sqrt{w_{vi}^Tw_{vi}+\varepsilon}}.
\end{split}
\end{equation}
With fixed $ W_v $, $ G $ is obtained by Eq. (\ref{G}). And with fixed $ G $, the solution of Eq. (\ref{W4}) can be obtained by solving
\begin{equation}\label{GETW}
\begin{split}
\min_{W_v^TW_v=1} \  Tr(W_{v}^T (X^{(v)})^T L_S X^{(v)} W_{v}) + \frac{\gamma}{\alpha_v}Tr(W_v^T G W_v).
\end{split}
\end{equation}
It is easy to know that the $ m_v$ column vectors of the optimal $ W_v $ are the $ m_v$ eigenvectors of $ ((X^{(v)})^T L_S X^{(v)} + \frac{\gamma}{\alpha_v}G) $, which correspond to the $ m_v$ smallest eigenvalues.
The details of deriving the solution of $ W_v $ is summarized in Algorithm 1, and the KKT condition is satisfied. In section \ref{subsec: Convergence Analysis of Alogrithm 1}, we will give its convergence proof.

\subsubsection{Update S with Fixed $ \alpha_v $, F and $ W_v $}
If $ \alpha_v $, F and $ W_v $ are fixed, we can rewrite the problem (\ref{2}) as the following problem:

\begin{equation}\label{s1}
\begin{split}
\min \limits_{S}\  &\sum_{v=1}^{V} ( \alpha_v Tr(W_{v}^T (X^{(v)})^T L_S X^{(v)} W_{v}) + \mu \sum_{i,j}s_{ij}^2+\\&2 \lambda Tr(F^T L_S F)\\
s.t. \ &\forall i, s_i^T \mathbf{1} = 1, 0 \leq s_{ij} \leq 1.
\end{split}
\end{equation}
In spectral analysis \cite{tutorialSP},
\begin{equation}\label{sf}
\begin{split}
\sum_{i,j} \lVert f_i-f_j \rVert_2^2 s_{ij} = 2Tr(F^T L_S F).
\end{split}
\end{equation}
So problem (\ref{s1}) can be rewritten as
\begin{equation}\label{s2}
\begin{split}
\min \limits_{S}\  &\sum_{i,j} ( \sum_{v=1}^{V} \alpha_v\lVert W_{v}^T x_i^{(v)}-W_{v}^T x_j^{(v)} \rVert_2^2 s_{ij} +  \mu s_{ij}^2)\\ &+\lambda \sum_{i,j} \lVert f_i-f_j \rVert_2^2 s_{ij}\\
s.t. \ &\forall i, s_i^T \mathbf{1} = 1, 0 \leq s_{ij} \leq 1.
\end{split}
\end{equation}
Because the row vectors of similarity matrix are independent with each other, problem (\ref{s2}) can be divided into $ n $ subproblems, and each subproblem aims to acquire the similarity vector of the corresponding sample. Here, we take the $ i $-th sample as an example.
\begin{equation}\label{s3}
\begin{split}
\min_{s_i^T \mathbf{1} = 1, 0 \leq s_{ij} \leq 1}&\sum_{j} ( \sum_{v=1}^{V} \alpha_v\lVert W_{v}^T x_i^{(v)}-W_{v}^T x_j^{(v)} \rVert_2^2 s_{ij} \\ &+  \mu s_{ij}^2) +\lambda \sum_{j} \lVert f_i-f_j \rVert_2^2 s_{ij}.\\
\end{split}
\end{equation}
Then, set $ p_{ij}=\sum_v \alpha_v \lVert W_v^T X_i^{(v)}-W_v^T X_j^{(v)}\rVert_{2}^2 $ , $ q_{ij}=\lVert f_i-f_j\rVert_2^2 $ and $ t_i \in \mathbb{R}^{n \times 1} $ whose element is $ t_{ij} = p_{ij} + \lambda q_{ij} $. Therefore, problem (\ref{s3}) can be transformed into 
\begin{equation}\label{s4}
\begin{split}
\min_{s_i^T \mathbf{1} = 1, 0 \leq s_{ij} \leq 1}\lVert s_i+ \frac{t_i}{2\mu} \rVert_2^2.
\end{split}
\end{equation}
The solution of this problem and the method to optimize parameter $ \mu $ will be shown later.

\subsubsection{Update F with Fixed $ \alpha_v $, $ W_v $ and S}
If $ S $, $ \alpha_v $ and $ W_v $ are fixed, problem (\ref{2}) becomes:
\begin{equation}\label{F}
\begin{split}
\min_{F^TF=I, F \in \mathbb{R}^{n \times c}}\lambda Tr(F^T L_S F).
\end{split}
\end{equation}
After acquiring the $ c $ smallest eigenvalues of $ L_S $, we can obtain the optimal solution of  $ F $ which is made up of $ c $ corresponding eigenvectors.

\subsubsection{Convergence Proof of Algorithm 1}
\label{subsec: Convergence Analysis of Alogrithm 1}
To prove the convergence of Algorithm 1, the lemma proposed by \cite{lemma} is needed,  which is described as follows.

\textbf{Lemma 1}: \emph{
	The following inequality holds for any positive real number $ u $ and $ v $.
	\begin{equation}\label{lemma}
	\begin{split}
	\sqrt{u}-\frac{u}{2\sqrt{v}} \leq \sqrt{v}-\frac{v}{2\sqrt{v}}.
	\end{split}
	\end{equation}
}

The converged solution of problem (\ref{W2}) can be derived by Algorithm \ref{alg:W_v}, and the convergence is proven by the following theorem.

\textbf{Theorem 1}: \emph{In Algorithm 1, updated $ W_v $ will decrease the objective value of problem (\ref{W2}) until converge.}

\textbf{Proof}: Let $ \widetilde{W}_v $ denotes the updated $ W_v $. Therefore,
\begin{equation}\label{prf1}
\begin{split}
&Tr( \widetilde{W}_v^T (X^{(v)})^T L_S X^{(v)} \widetilde{W}_v ) +\frac{\gamma}{\alpha_v} Tr(\widetilde{W}_v^T  G \widetilde{W}_v ) \\
\leq&Tr( {W_v}^T (X^{(v)})^T L_S X^{(v)} {W_v} ) +\frac{\gamma}{\alpha_v} Tr({W_v}^T G {W_v} ).
\end{split}
\end{equation}
Then, by adding $ \frac{\gamma}{\alpha_v} \sum_i\frac{\varepsilon}{2\sqrt{w_{vi}^Tw_{vi}+\varepsilon}} $ to the both sides and substituting the definition of $ G $ shown in Eq. (\ref{G}), the inequality (\ref{prf1}) changes to

\begin{equation}\label{prf2}
\begin{split}
&Tr( \widetilde{W}_v^T (X^{(v)})^T L_S X^{(v)} \widetilde{W}_v ) +\frac{\gamma}{\alpha_v} \sum_i\frac{\widetilde{w}_{vi}^T \widetilde{w}_{vi}+\varepsilon}{2\sqrt{w_{vi}^Tw_{vi}+\varepsilon}} \\
\leq&Tr( {W_v}^T (X^{(v)})^T L_S X^{(v)} {W_v} ) + \frac{\gamma}{\alpha_v} \sum_i\frac{w_{vi}^T w_{vi}+\varepsilon}{2\sqrt{w_{vi}^Tw_{vi}+\varepsilon}}. 
\end{split}
\end{equation}
According to Lemma 1, it is easy to know
\begin{equation}\label{prf3}
\begin{split}
&\frac{\gamma}{\alpha_v} \sqrt{\widetilde{w}_{vi}^T \widetilde{w}_{vi}+\varepsilon} - \frac{\gamma}{\alpha_v} \sum_i\frac{\widetilde{w}_{vi}^T \widetilde{w}_{vi}+\varepsilon}{2\sqrt{w_{vi}^Tw_{vi}+\varepsilon}} \\
\leq&\frac{\gamma}{\alpha_v}  \sqrt{w_{vi}^T w_{vi}+\varepsilon} -\frac{\gamma}{\alpha_v}\sum_i\frac{w_{vi}^T w_{vi}+\varepsilon}{2\sqrt{w_{vi}^Tw_{vi}+\varepsilon}}. 
\end{split}
\end{equation}
Finally, by summing the inequality (\ref{prf2}) and inequality (\ref{prf3}), we get the following inequality and complete the proof.
\begin{equation}\label{prf4}
\begin{split}
&Tr( \widetilde{W}_v^T (X^{(v)})^T L_S X^{(v)} \widetilde{W}_v ) +\frac{\gamma}{\alpha_v} \sqrt{\widetilde{w}_{vi}^T \widetilde{w}_{vi}+\varepsilon} \\
\leq
&Tr( {W_v}^T (X^{(v)})^T L_S X^{(v)} {W_v} ) +\frac{\gamma}{\alpha_v} \sqrt{w_{vi}^T w_{vi}+\varepsilon}.
\end{split}
\end{equation}

\begin{algorithm}
	\setcounter{algorithm}{0}
	\caption{Algorithm to get projection matrices $ W_v $}
	\label{alg:W_v}
	
	{\bf Input:}
	The  data matrix $\{ X^{(1)}, X^{(2)}, \cdots, X^{(V)} \}$, $ X^{(v)}\in \mathbb {R}^{d_v \times n} $, regularization parameter $ \gamma $, the view weights $ \alpha_v $, projection dimension $ m_v $, laplacian matrix $ L_S \in \mathbb{R}^{n \times n} $.
	\vspace*{0.08in}
	Initialize $ G\in \mathbb{R}^{d_v \times d_v} $ as $ G=I $.
	
	For each view:
	
	{\bf Repeat}
	
	1. Get the solution $ W_v $ of problem (\ref{GETW}).
	
	2. Update $ G $ via Eq. (\ref{G}).
	
	{\bf Until} converge
	\vspace*{0.08in}	
	
	{\bf Output:}
	
	Projection matrices $ \{W_v \in \mathbb{R}^{d_v \times m_v}\}_{v=1}^V $
	
\end{algorithm}

\begin{algorithm}
	\setcounter{algorithm}{1}
	\caption{Algorithm to solve MFSGL}
	\label{alg:final}
	
	{\bf Input:}
	The  multi-view dataset $\{ X^{(1)}, X^{(2)}, \cdots, X^{(V)} \}$, $ X^{(v)}\in \mathbb {R}^{d_v \times n} $,  regularization parameter $ \gamma $, number of selected features $ s $, number of clusters $ c $, projection dimension $ m_v  $, a large enough $ \lambda $.
	
	Initialize $ \{\alpha_v\}_{v=1}^V = \frac{1}{V}$.
	
	Initialize $ S $ by solving
	\par 
	\begin{center}
		$\min \limits_{S} \sum_{i,j} ( \sum_{v=1}^{V} \alpha_v\lVert x_i^{(v)}-x_j^{(v)} \rVert_2^2 s_{ij} + \mu s_{ij}^2$)\par 
		$ s.t. \forall i, s_i^T \mathbf{1} = 1, 0 \leq s_{ij} \leq 1 $.
	\end{center}
	
	\vspace*{0.08in}
	{\bf Repeat}
	
	1. Obtain the diagonal matrix $ D $ and $ L_S $ according to $ D_{ii}=\sum_j \frac{s_{ij}+s_{ji}}{2} $ and $ L_S=D-\frac{S^T+S}{2} $ respectively.
	
	2. Update $W_v$ for each view by using Algorithm 1.
	
	3. Obtain $  F $ by solving problem (\ref{F}).
	
	4. Obtain each $ s_i $ via solving problem (\ref{s4}).
	
	5. Update the parameter $ \mu $ according to Eq. (\ref{a4}).
	
	6. Update the view weights $ \alpha_v $ for each view according to Eq. (\ref{i2}).
	
	{\bf Until} converge
	\vspace*{0.08in}	
	
	{\bf Output:}
	Features sorted by $ \Vert w_{vi} \rVert_2 $ in descending order. The final feature subset which is made up of the top $ s $ features.
	
\end{algorithm}

\begin{table*}[htbp] 
	\centering
	\caption{\label{tab:test}Details of the multi-view datasets used in our experiments.} 
	\setlength{\tabcolsep}{9mm}{
		\begin{tabular}{ccccc}
			\hline	
			view&\textit{Outdoor-Scene}&\textit{Caltech101-7}&\textit{NUS-WIDE-OBJ}&\textit{Handwritten}\\
			\hline
			1&GIST (512)&GABOR (48)&CH (64)&FAC (216)\\
			2&CM (432)&WM (40)&CMT (225)&PIX (240)\\
			3&HOG (256)&CENTRIST (254)&CORR (144)&ZER (47)\\
			4&LBP (48)&HOG (1984)&EDH (73)&MOR (6)\\
			5&-&GIST (512)&WT (128)&KAR (64)\\
			6&-&LBP (928)&-&FOU (76)\\
			\hline
			Number of features&1622&3766&634&649\\
			Number of samples&210&8677&3000&2000\\
			Classes&7&7&25&10\\
			\hline 
	\end{tabular} }
	\label{dataset}
\end{table*}
\begin{figure*}[htbp]
	\centering
	\begin{minipage}[b]{0.329\linewidth}
		\centering
		\centerline{\includegraphics[width=5cm]{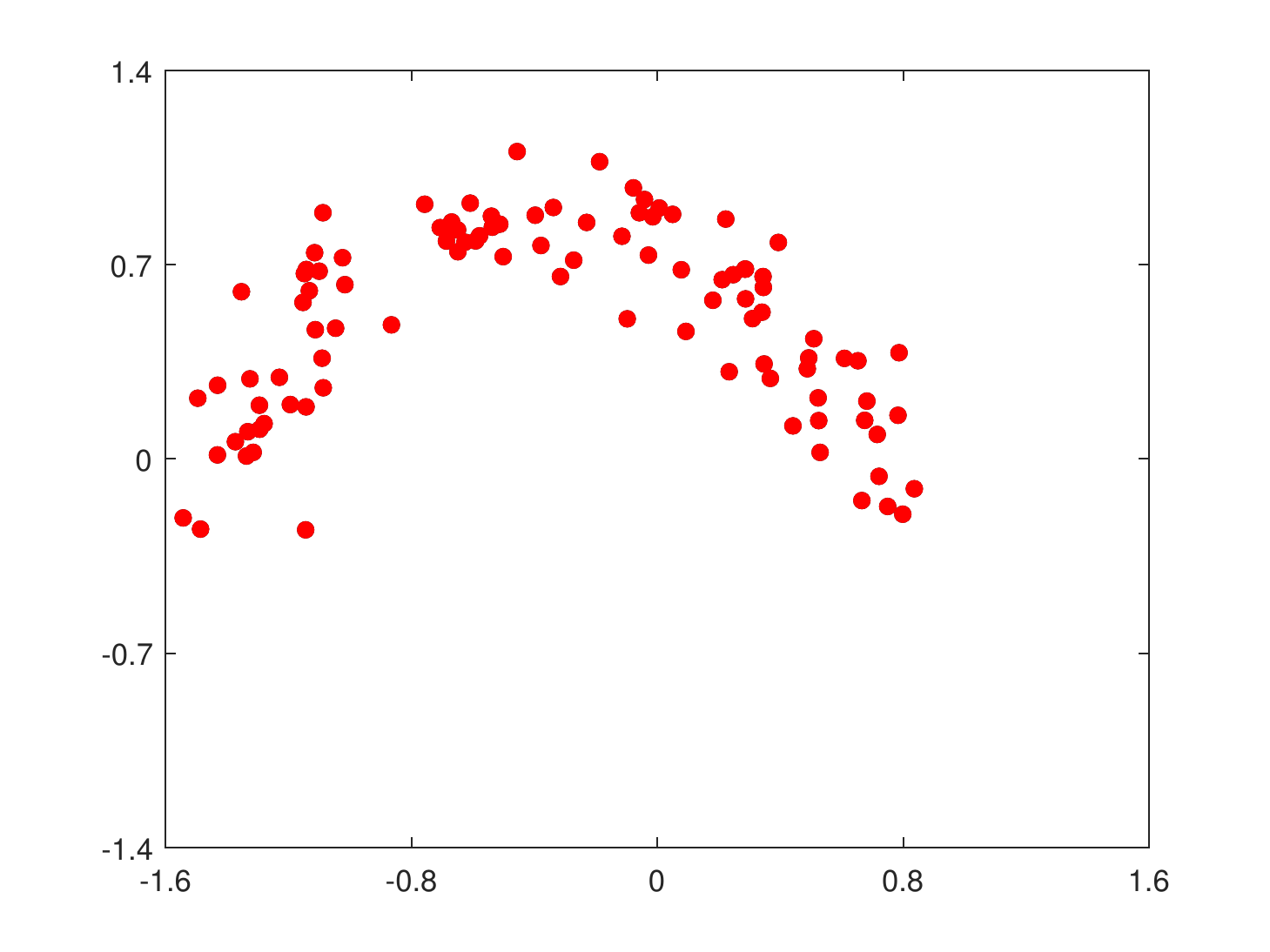}}
		\centerline{View 1}\medskip
	\end{minipage}
	\hfill	
	\begin{minipage}[b]{0.329\linewidth}
		\centering
		\centerline{\includegraphics[width=5cm]{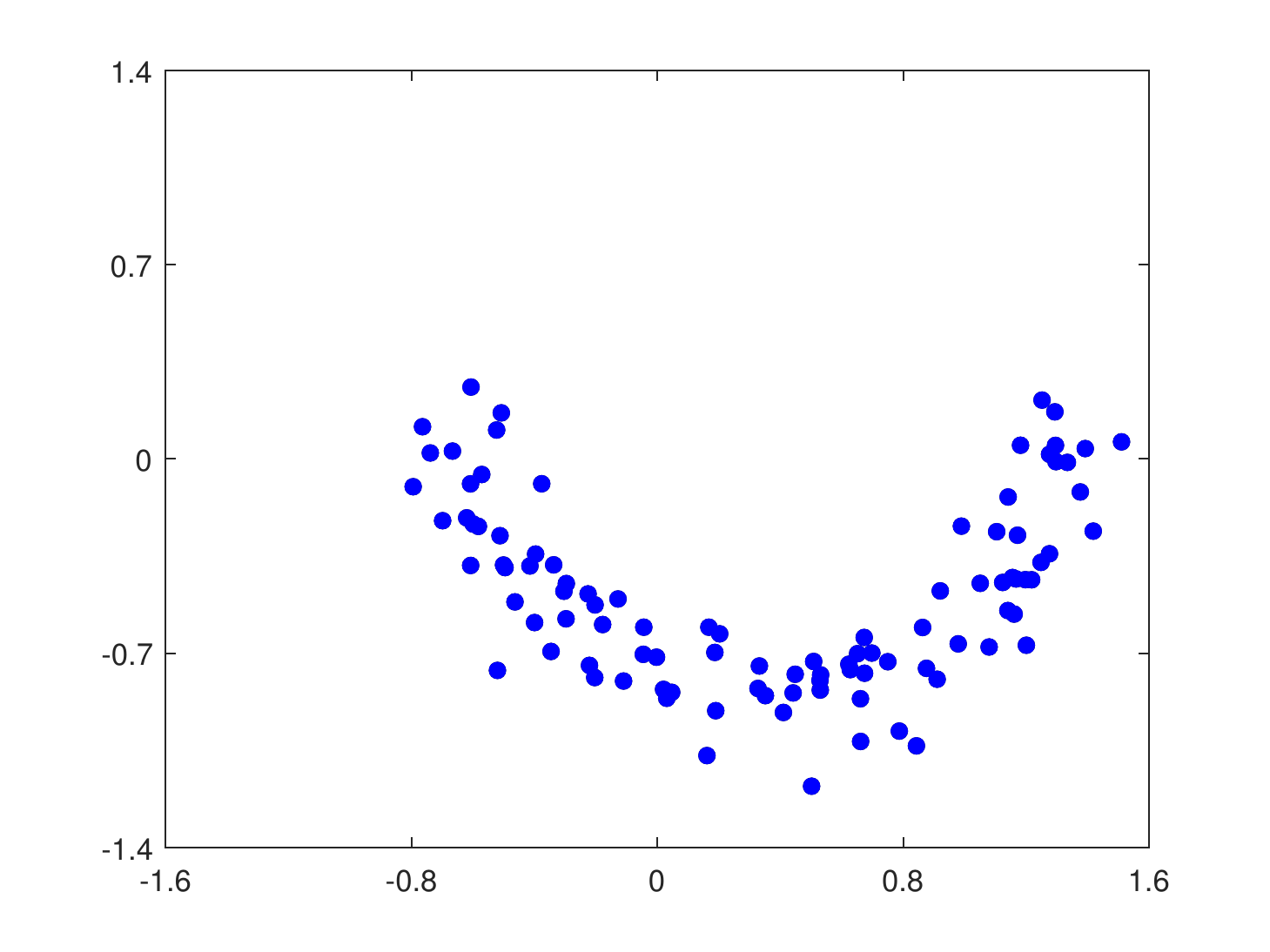}}
		\centerline{View 2}\medskip
	\end{minipage}
	\hfill	
	\begin{minipage}[b]{0.329\linewidth}
		\centering
		\centerline{\includegraphics[width=5cm]{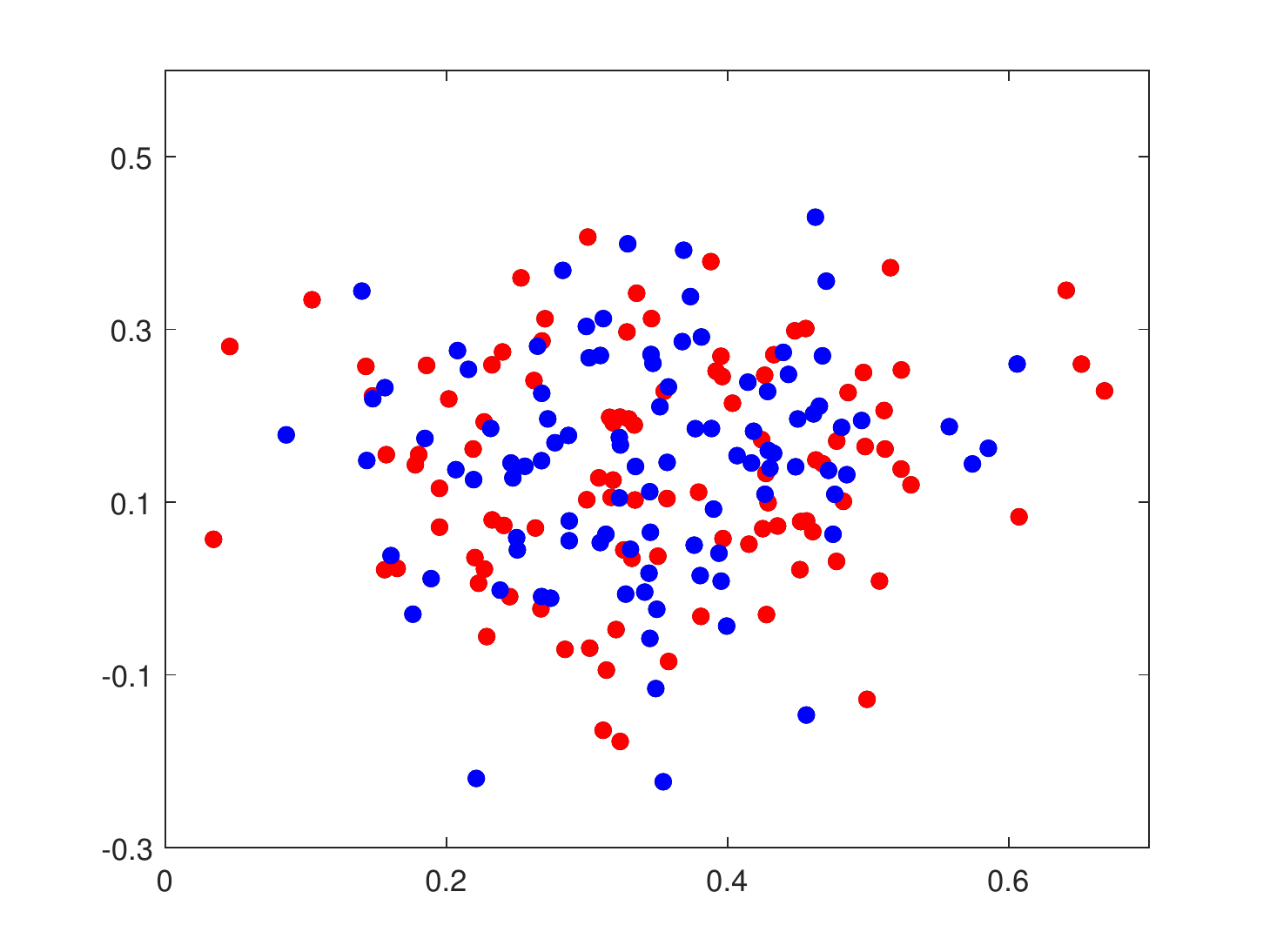}}
		\centerline{Noise View}\medskip
	\end{minipage}
	\caption{The synthetic multi-view two-moon dataset.}
	\label{v12}
\end{figure*}
\begin{figure*}[htbp]
	\centering
	\begin{minipage}[b]{0.329\linewidth}
		\centering
		\centerline{\includegraphics[width=5cm]{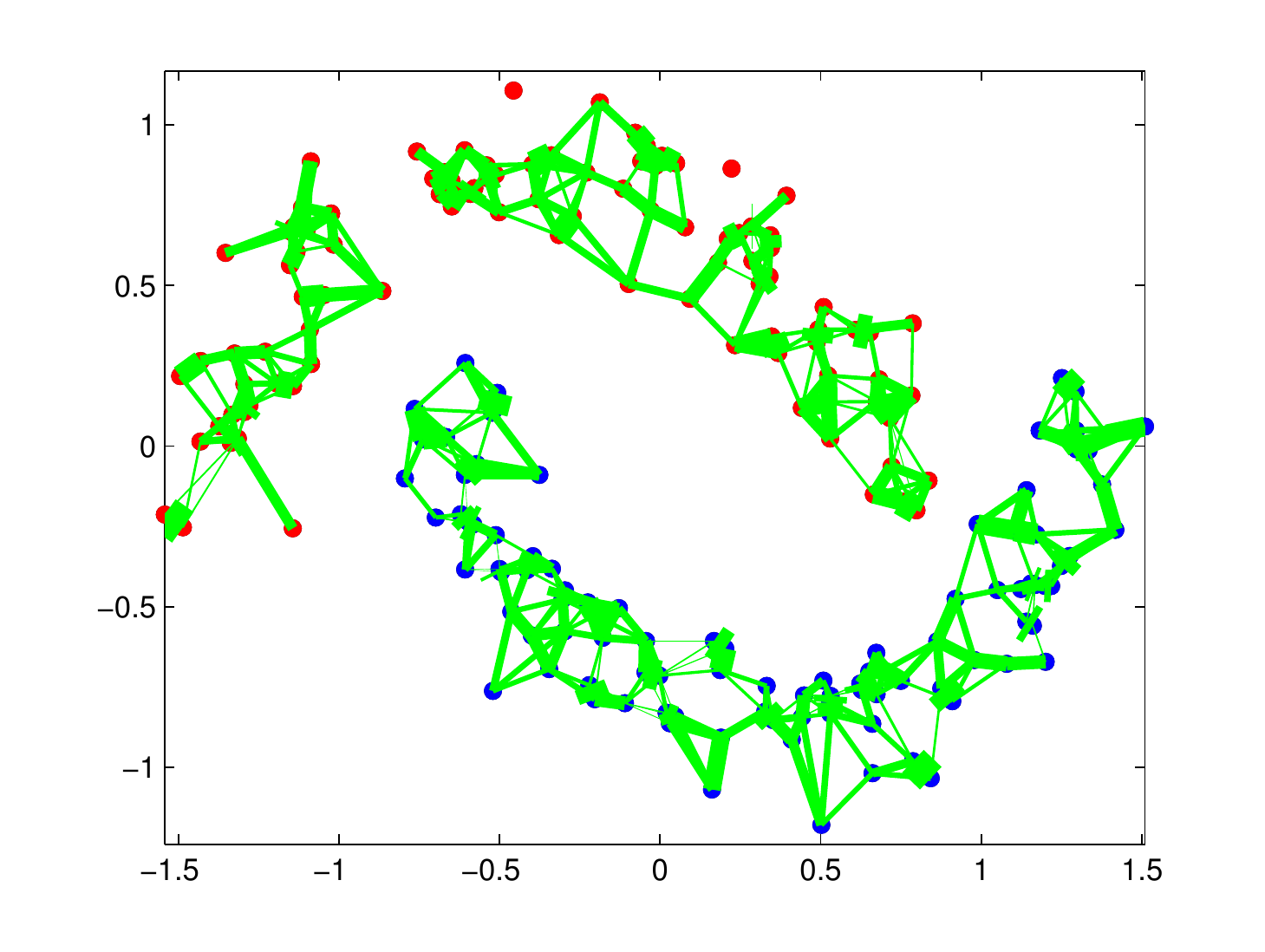}}
		\centerline{ASVW, $k$ = 5}\medskip
	\end{minipage}
	\begin{minipage}[b]{0.329\linewidth}
		\centering
		\centerline{\includegraphics[width=5cm]{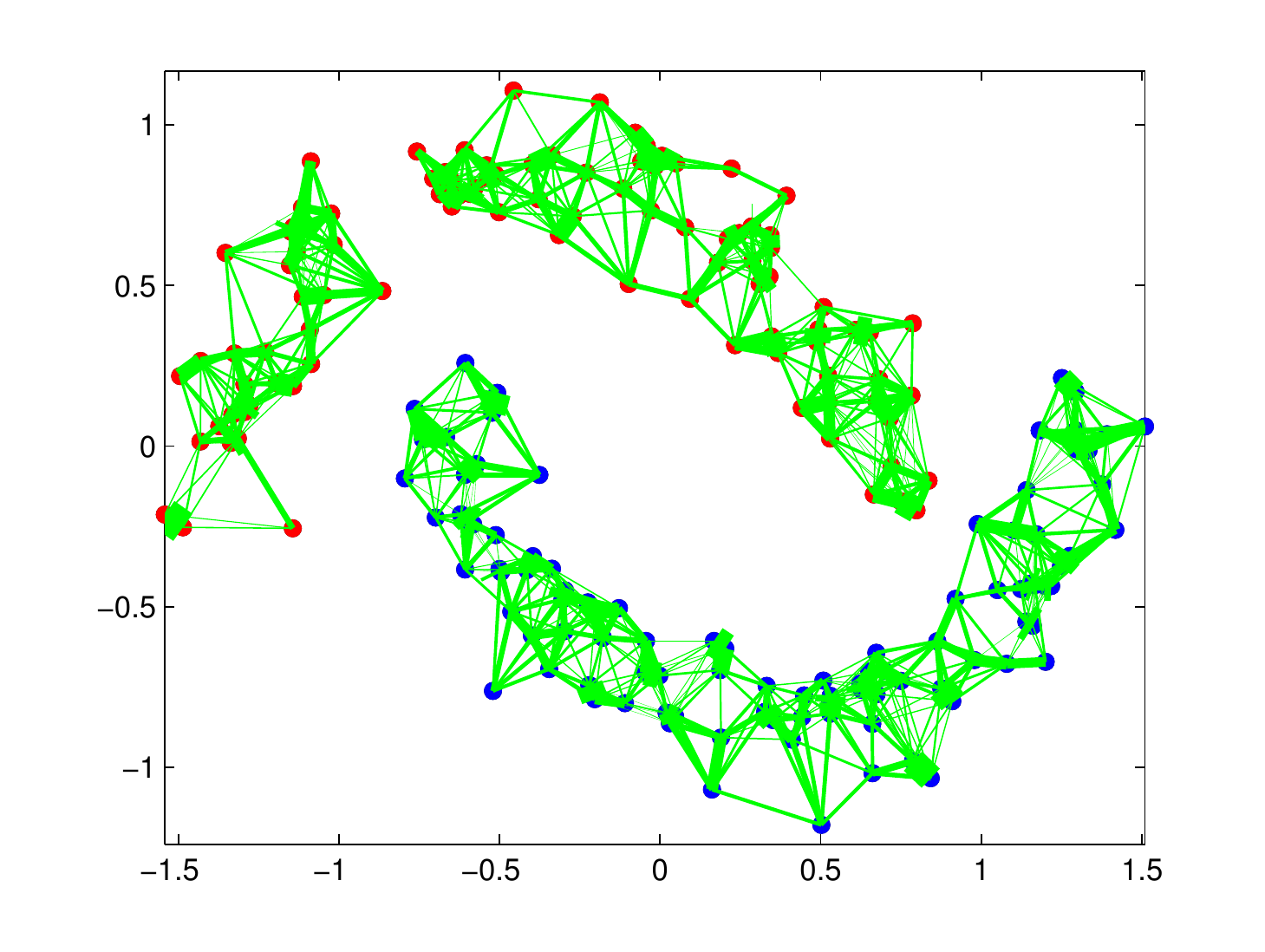}}
		\centerline{ASVW, $k$ = 10}\medskip
	\end{minipage}
	\begin{minipage}[b]{0.329\linewidth}
		\centering
		\centerline{\includegraphics[width=5cm]{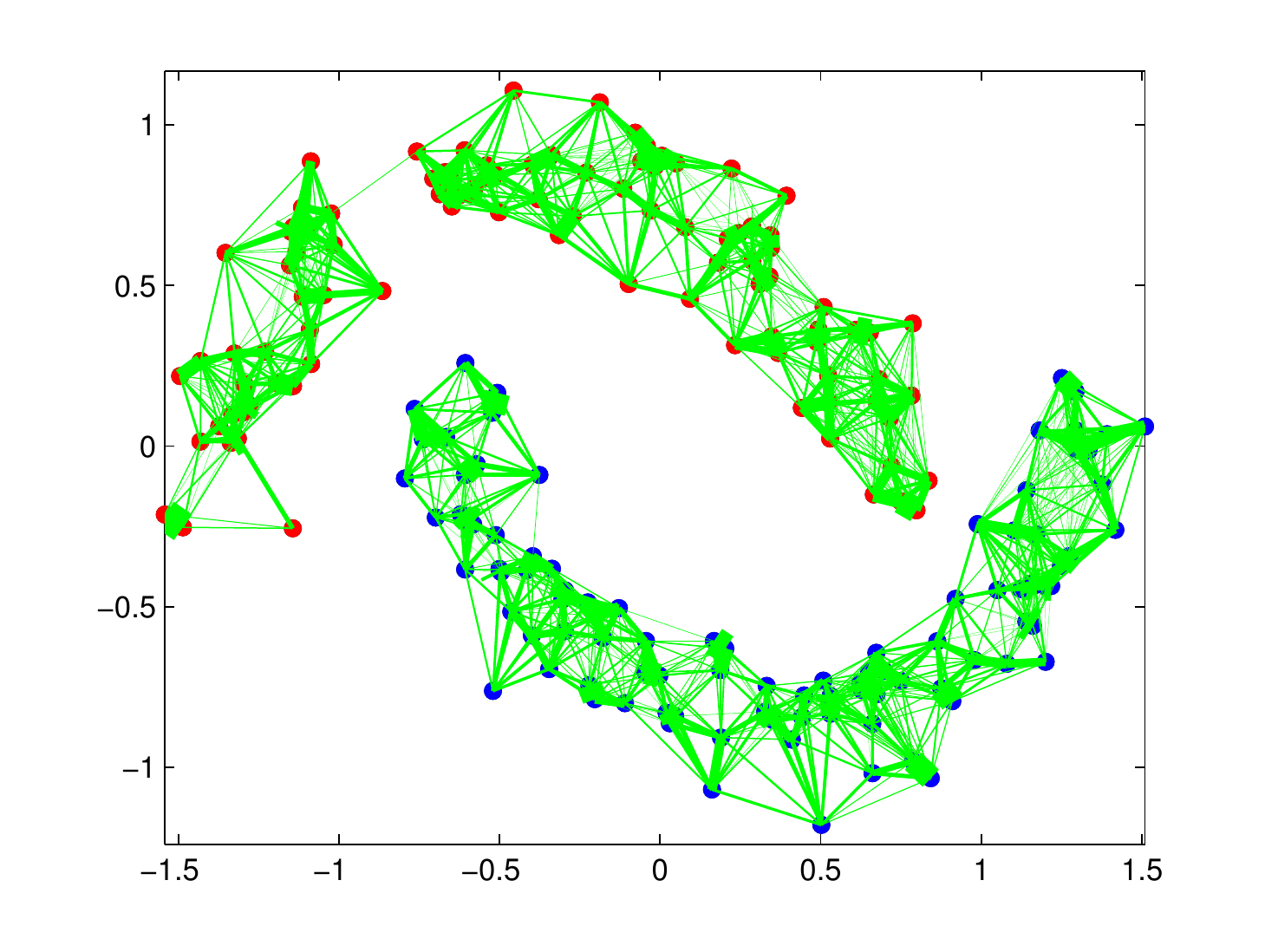}}
		\centerline{ASVW, $k$ = 15}\medskip
	\end{minipage}
	\\
	\begin{minipage}[b]{0.329\linewidth}
		\centering
		\centerline{\includegraphics[width=5cm]{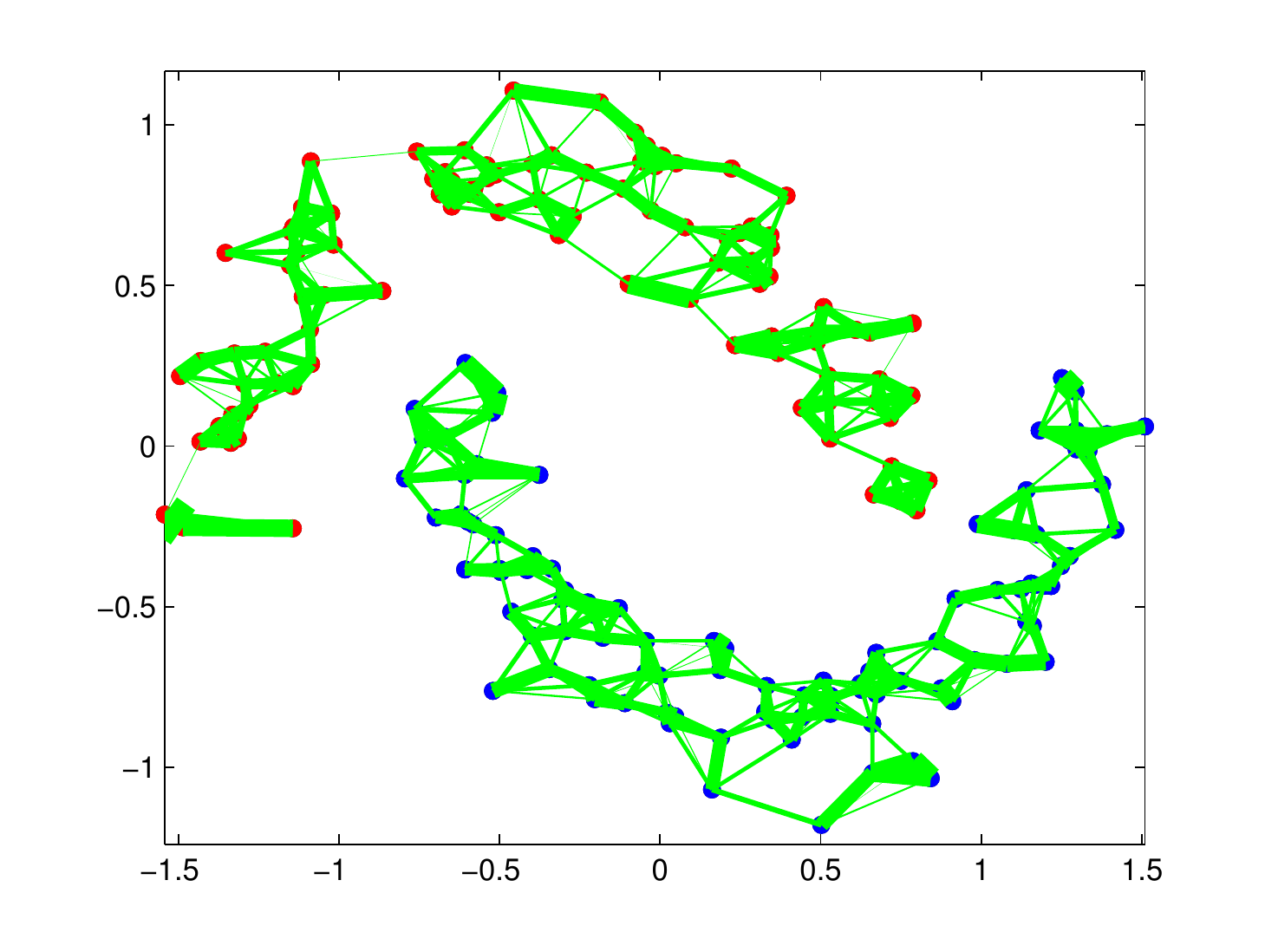}}
		\centerline{MFSGL, $k$ = 5}\medskip
	\end{minipage}
	\begin{minipage}[b]{0.329\linewidth}
		\centering
		\centerline{\includegraphics[width=5cm]{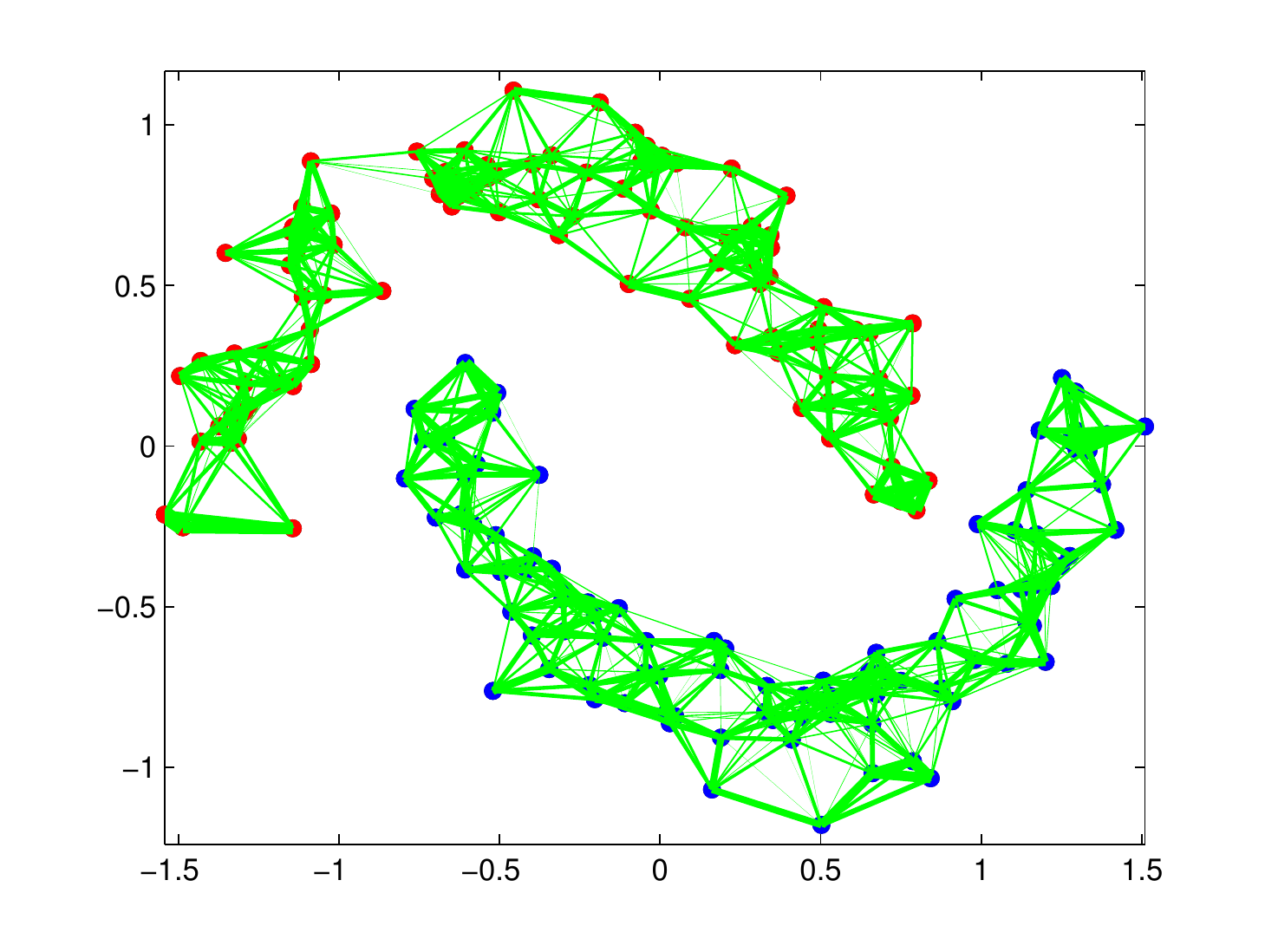}}
		\centerline{MFSGL, $k$ = 10}\medskip
	\end{minipage}
	\begin{minipage}[b]{0.329\linewidth}
		\centering
		\centerline{\includegraphics[width=5cm]{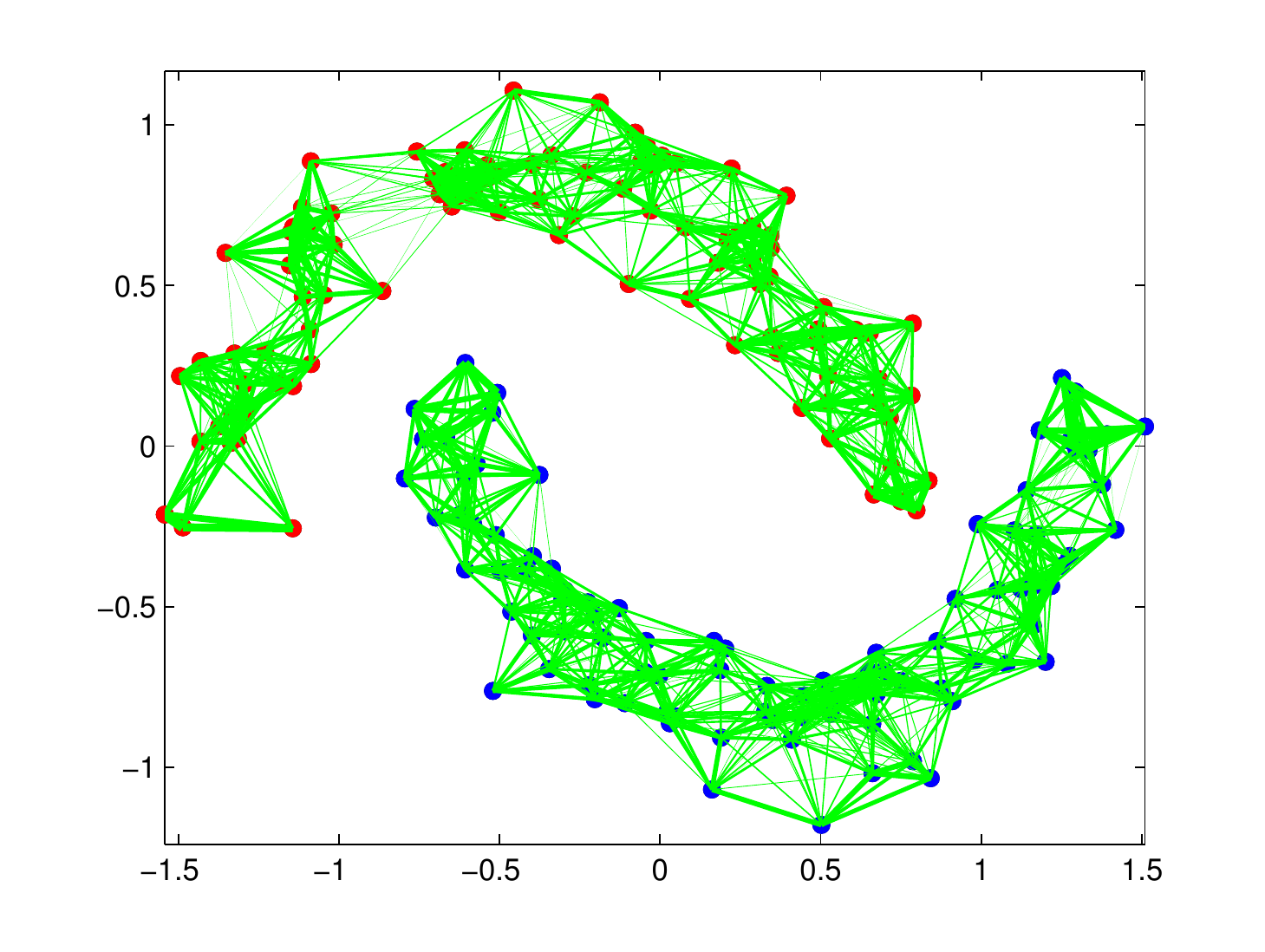}}
		\centerline{MFSGL, $k$ = 15}\medskip
	\end{minipage}
	
	\caption{The graphs learned by MFSGL and ASVW on pure two-moon dataset.}
	\label{pure}
\end{figure*}

\subsubsection{Determination of $\mu$}
It has been noted that the parameter $ \mu $ is used to avoid the trivial solution. Let us consider two extreme conditions of $ \mu $. When $ \mu=0 $, the similarity between the closet two samples is 1, and the others are 0. When $ \mu=\infty $, all similarities should equal to $  \frac{1}{n} $. Therefore, $ \mu $ is crucial to the number of sample neighbors.
\par 
If the neighbor number of a sample is set to be $ k $, here we derive the optimal $ \mu $ to ensure that there are $ k $ non-zero elements in most $ s_i $. 

$ \theta $ and $ \varphi_i $ are denoted as the Lagrangian multipliers, and the Lagrangian function of problem (\ref{s4}) can be constructed as
\begin{equation}\label{a1}
\begin{split}
\mathbf{\mathcal{L}}(s_i,\theta,\varphi_i) = 
\frac{1}{2}\lVert s_i+ \frac{t_i}{2\mu_i} \rVert_2^2-\theta (s_i^T  \mathbf{1}-1)- \varphi_i^T s_i.
\end{split}
\end{equation}
According to the KKT condition, the solution of this problem is obtained.
\begin{equation}\label{a2}
\begin{split}
s_{ij}=(-\frac{t_i}{2\mu_i}+\theta)_+.
\end{split}
\end{equation}
Here, $ \theta = \frac{1}{k} + \frac{1}{2k \mu_i}\sum_{j=1}^{k}t_{ij} $ \cite{16}.
$ s_i $ should have $ k $ non-zero elements exactly, that is to say, $ s_{i,k+1} \leq 0 < s_{i,k} $. Therefore, the $ \mu_i $ should satisfy the following property:
\begin{equation}\label{a3}
\begin{split}
\frac{k}{2}t_{i,k}-\frac{1}{2}\sum_{j=1}^{k}t_{ij}< \mu_i < \frac{k}{2} t_{i,k+1}- \frac{1}{2}\sum_{j=1}^{k}t_{ij},
\end{split}
\end{equation}
and we can set a good enough $ \mu $ as
\begin{equation}\label{a4}
\begin{split}
\mu = \frac{1}{n} \sum_{i=1}^{n}\mu_i=\frac{1}{n} \sum_{i=1}^{n} (\frac{k}{2} t_{i,k+1}- \frac{1}{2}\sum_{j=1}^{k}t_{ij}),
\end{split}
\end{equation}
 where $ t_{i1}, t_{i2}, \cdots, t_{in} $ are sorted in ascending order.
\par 
In summary, we give an alternative iteration method to optimize the objective function of MFSGL. $ \alpha_v $ can be modified according to Eq. (\ref{i2}). $ S, W_v$ and $ F $ can be modified by tackling problem (\ref{2}). The details are summarized in Algorithm \ref{alg:final}.

\section{EXPERIMENTS}
\label{sec:EXPERIMENTS}
Here, the effectiveness of MFSGL is demonstrated on both synthetic datasets and real-world datasets.

\subsection{Experiments on the Synthetic Datasets}
In this part, we first randomly generate the two-moon dataset to demonstrate the superiority of the graph learning strategy of MFSGL. In this dataset, two clusters of data samples are scattered in the two-moon space and there are  100 data points in each cluster. As shown in Fig. \ref{v12}, the two clusters are labeled in red and blue respectively, and they are separated into two independent views. In addition, a noise view is also generated. An ideal multi-view similarity graph learning method should integrate the different views, and distinguish the two clusters exactly. View 1 and View 2 make up the pure two-moon dataset. View 1, View 2 and Noise View make up the noisy two-moon dataset. The experiments are conducted in the two datasets respectively. The parameter $p$ of MFSGL is set as 1.

Firstly, on the pure two-moon dataset, we fix the neighbor number $k$ as 5, 10 and 15, and show the learned similarity graphs of MFSGL and ASVW in Fig. \ref{pure}. When $k$ = 5, 10, some pairs of data samples contained in the same cluster are not linked in the learned graph of ASVW. That is to say, ASVW fails to separate data samples into two clusters. Whereas, in the learned graph of MFSGL, the data samples are divided into two clusters successfully. When $k$ = 15, although ASVW gives the correct result, MFSGL shows more strength. 
Both ASVW and MFSGL learn a common similarity graph of all views during feature selection. In MFSGL, a rank constraint is added to the similarity matrix, which brings that the number of connected components in the learned graph equals to the number of clusters, and each connected component corresponds to one cluster. Therefore, MFSGL is able to capture the relationship between data samples accurately from multiple views.

\begin{figure}[h]
	\centering
	\begin{minipage}[b]{0.49\linewidth}
		\centering
		\centerline{\includegraphics[width=4.4cm]{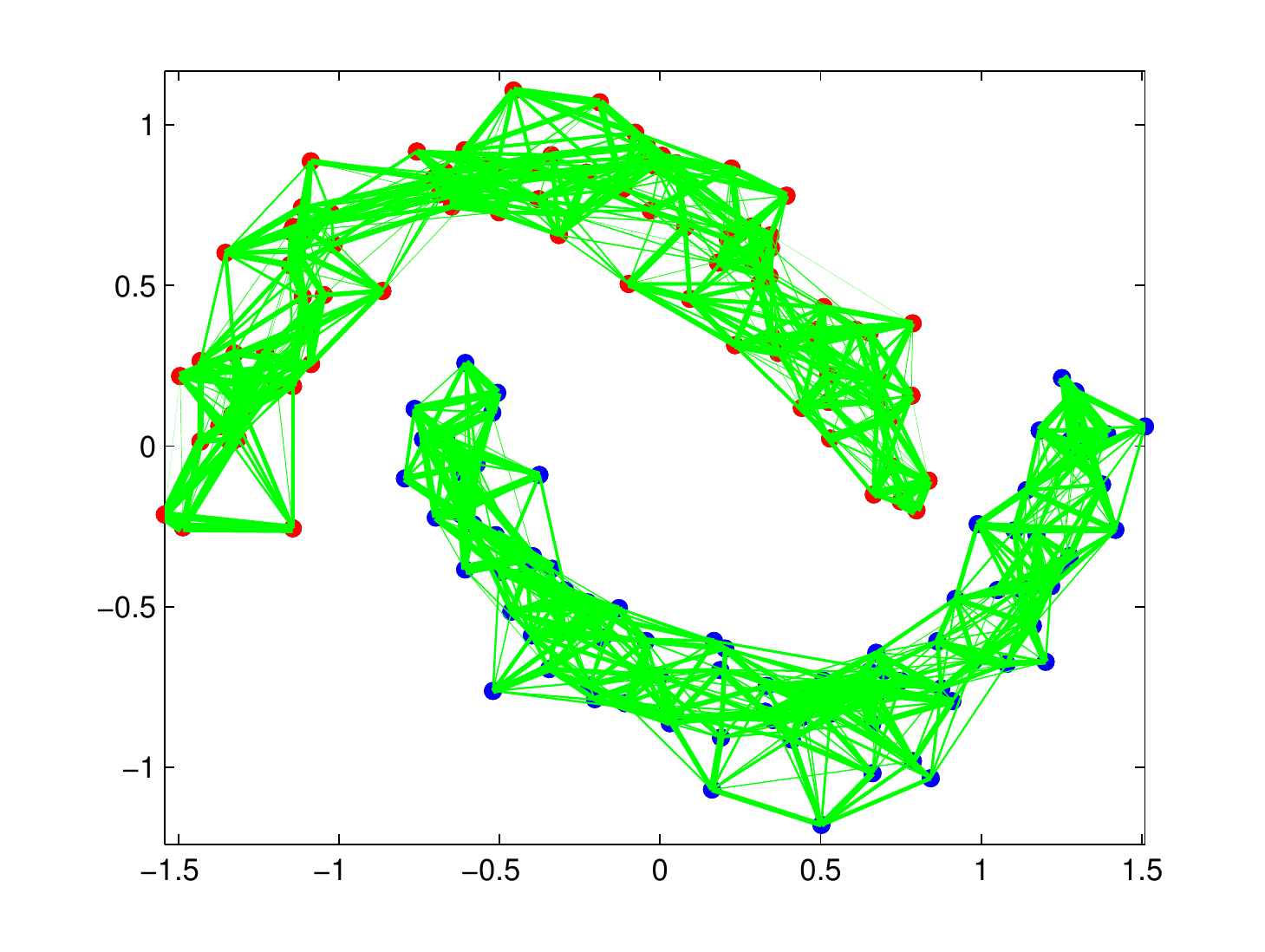}}
		\centerline{MFSGL}\medskip
	\end{minipage}
	\begin{minipage}[b]{0.49\linewidth}
		\centering
		\centerline{\includegraphics[width=4.4cm]{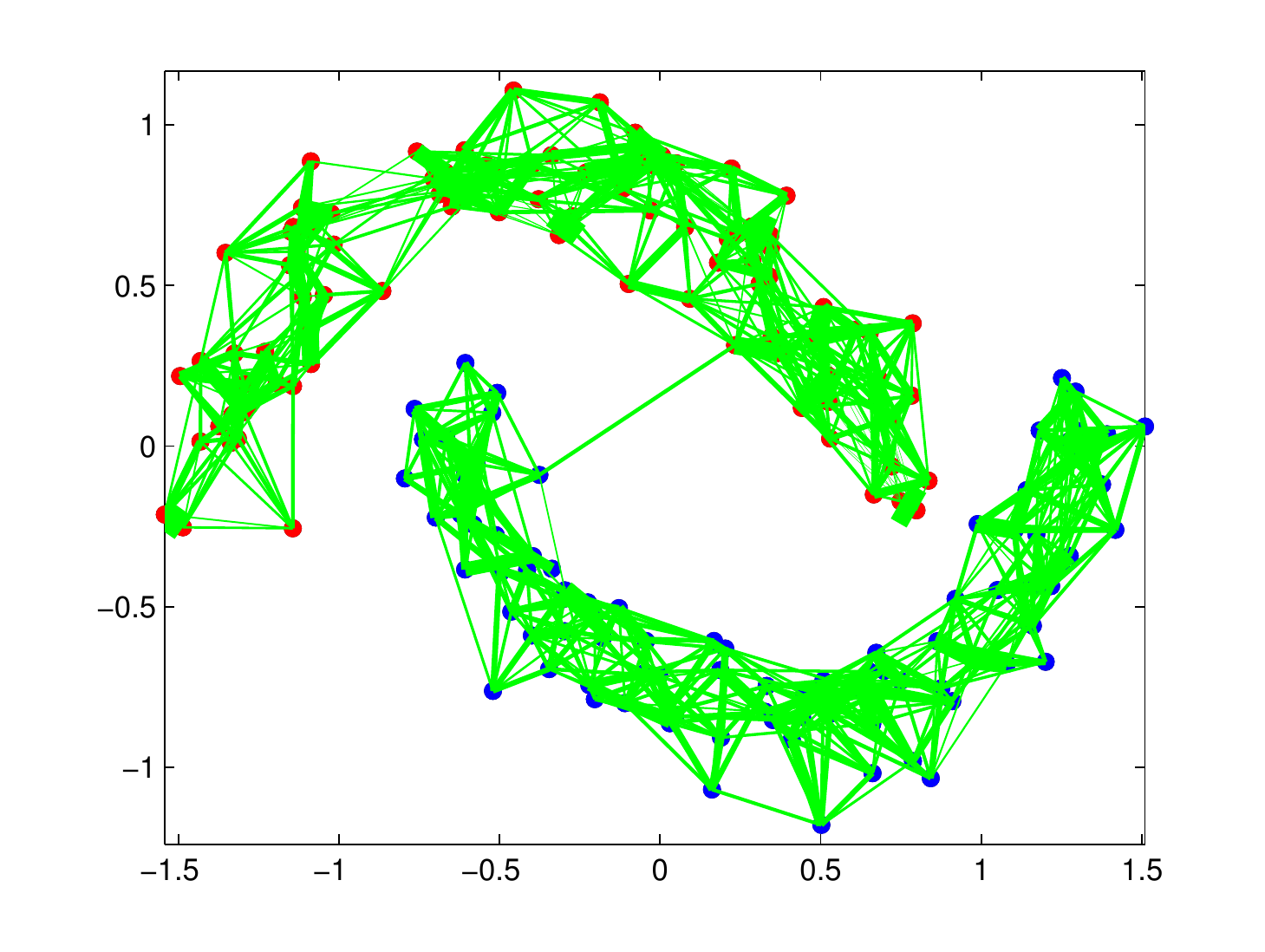}}
		\centerline{ASVW}\medskip
	\end{minipage}
	
	\begin{minipage}[b]{0.49\linewidth}
		\centering
		\centerline{\includegraphics[width=4.4cm]{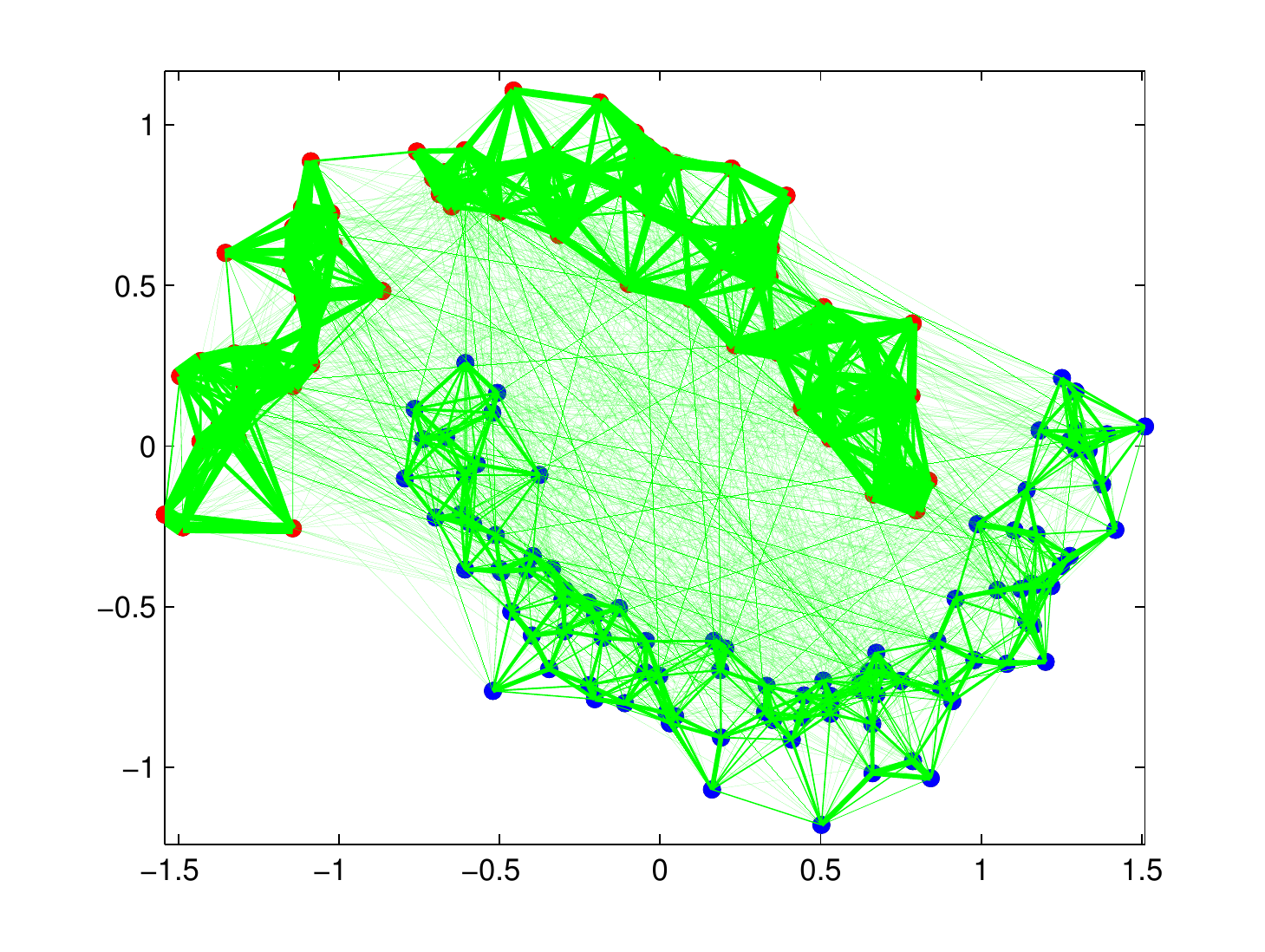}}
		\centerline{AMGL}\medskip
	\end{minipage}	
	\begin{minipage}[b]{0.49\linewidth}
		\centering
		\centerline{\includegraphics[width=4.4cm]{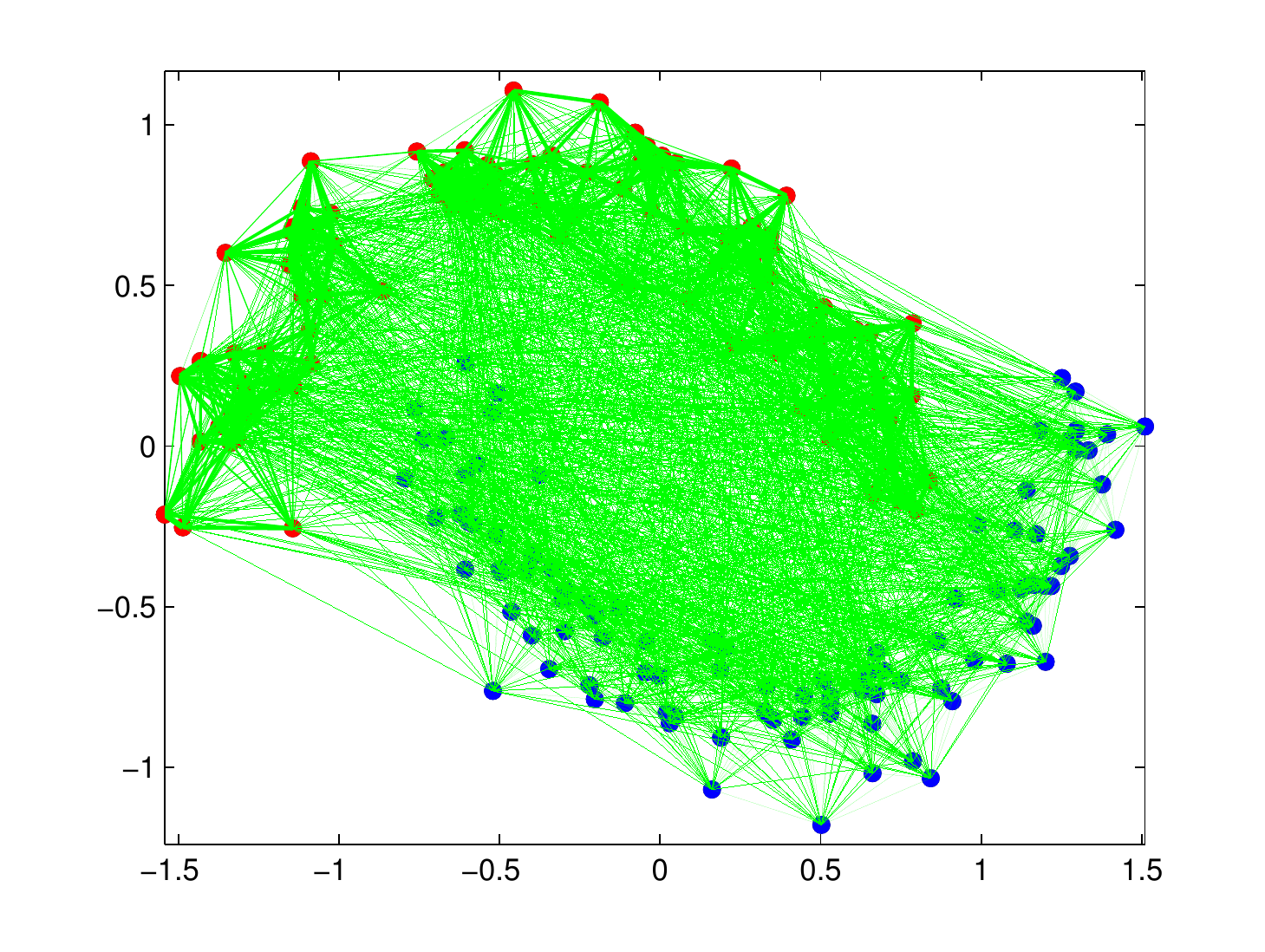}}
		\centerline{AASC}\medskip
	\end{minipage}	
	\caption{The graphs learned on noisy two-moon dataset.}
	\label{noisy}
\end{figure}

Secondly, to further verify the efficiency of proposed multi-view similarity graph learning strategy, another experiment is conducted on the noisy two-moon dataset. The neighbor number $k$ of all methods are set as 10. Affinity Aggregation Spectral Clustering (AASC) \cite{AASC} and Auto-weighted Multiple Graph Learning (AMGL) \cite{lijing} both pre-define the similarity graph of each view, and connect them by the learned view weights. The result in Fig. \ref{noisy} shows that AASC, AMGL and ASVW all fails to divide the data samples into correct clusters. Obviously,
MFSGL can learn the data structural information from different views precisely, and is robust to noise as well.

\begin{figure}[h]
	\centering
	\begin{minipage}[b]{0.49\linewidth}
		\centering
		\centerline{\includegraphics[width=4.4cm]{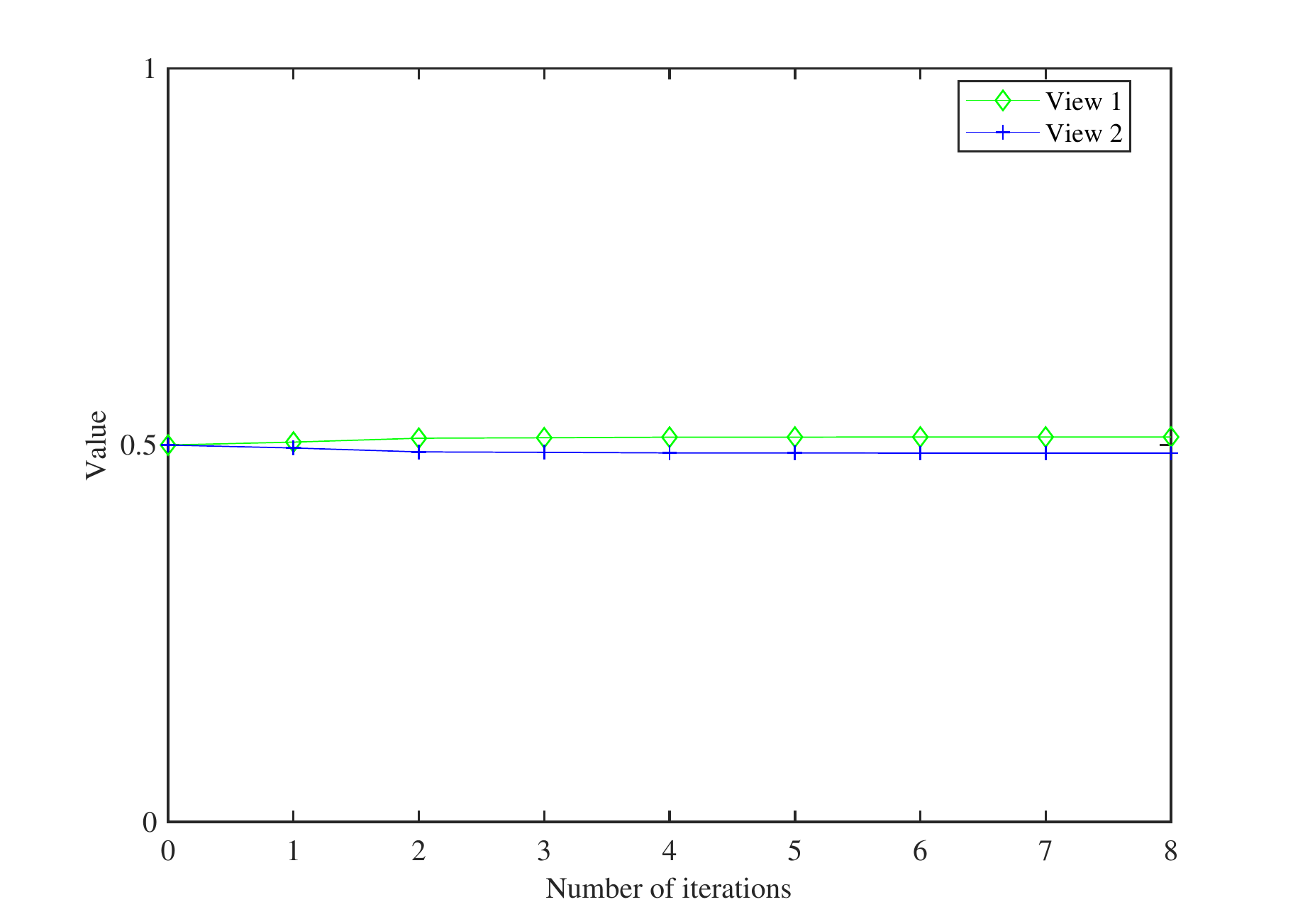}}
		\centerline{The pure two-moon dataset}\medskip
	\end{minipage}
	\hfill	
	\begin{minipage}[b]{0.49\linewidth}
		\centering
		\centerline{\includegraphics[width=4.4cm]{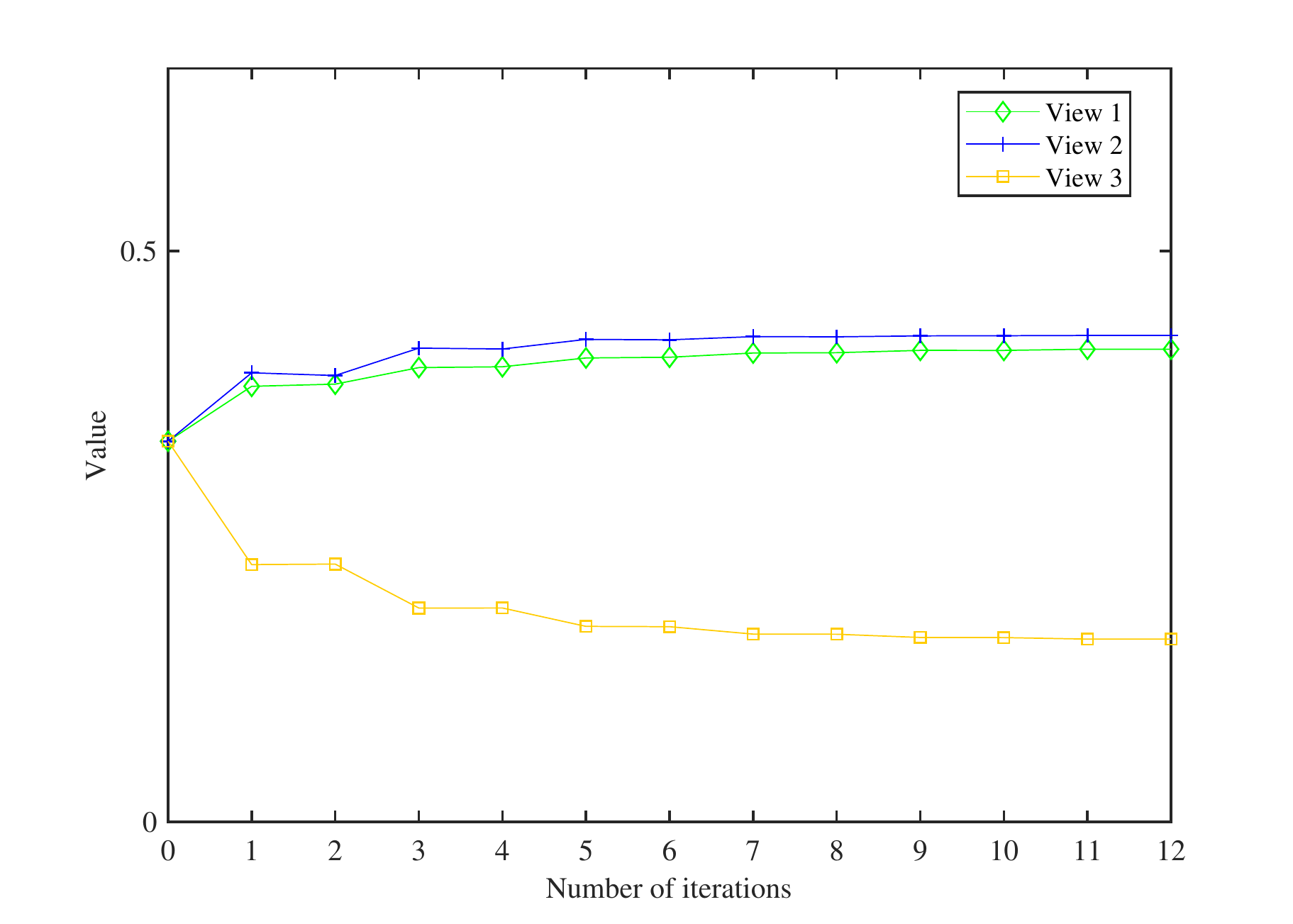}}
		\centerline{The noisy two-moon dataset}\medskip
	\end{minipage}
	\caption{The iterative curve of the view weights on two datasets.}
	\label{Weight}
\end{figure}

Moreover, the iterative curve of the view weights learned by MFSGL are demonstrated in Fig. \ref{Weight}. The presented results have been normalized. On the pure two-moon dataset, the two views contain complementary information, and the learned view weights are approximately equal to 0.5. On the noisy two-moon dataset, the weights of first two views increase while the third view weight decreases, which complies the assumption that noisy view should be with small weight. Therefore, MFSGL is able to assign an appropriate weight for each view according to their significance.

\subsection{Experiments on the Real-world Datasets}
In this part, MFSGL is compared with several state-of-the-art unsupervised feature selection approaches on four real-world datasets to demonstrate its effectiveness.
\subsubsection{Real-world Datasets Description}
In our experiments, four benchmark multi-view datasets are used, including Outdoor-Scene \cite{scene}, Caltech101-7 \cite{27}, NUS-WIDE-OBJ \cite{28} and Handwritten numerals \cite{Handwritten}. In each dataset, heterogeneous features are extracted as different views. The details of them are summarized in Table \ref{dataset}.
\begin{itemize}
	\item \textit{Outdoor-Scene}. This dataset includes 2688 images which belong to the following categories: mountain, open country, coast, forest, inside city, highways, street and tall buildings. For each image, we extract four kinds of visual features as different data views.
	\item \textit{Caltech101-7}. This dataset is commonly used for object recognition. Caltech101 includes 8677 images from 101 classes. Following \cite{ASVW}, we choose 1474 pictures totally of the widely used 7 categories, i.e., Faces, Motorbikes, Dolla-Bill, Garfield, Snoopy, Stop-Sign and Windsor-Chair. Different visual features are also extracted as different data views.
	\item \textit{NUS-WIDE-OBJ}. This dataset consists of 30,000 images of 31 categories totally. In our experiment, we select 25 categories with a total of 3,000 images and extract five kinds of features to represent each image.
	\item \textit{Handwritten numerals}. This dataset contains 2000 digital images from 0 to 9, and each category contains 200 images. Six different features are extracted in this dataset.
\end{itemize}

\subsubsection{Comparison Scheme}
Several representative single-view feature selection methods including LabScor, MCFS , RSFS and SOGFS are employed to show the effectiveness of MFSGL, and they take the samples represented by the connected features of all views as input. ASVW, RMFS and DEKM \cite{DEKM} are the multi-view feature selection methods. In addition, the connected features of all views is used to perform K-means as the baseline. For the selected feature subset of different size, we perform K-means from the same starting points to make the experiments fair enough. The parameter $k$ in MFSGL is searched from 5 to 15 with the step size 5. The parameter $\gamma$ in MFSGL is searched in logarithm form,  i.e., $log_{10}\gamma$ is searched from -2 to 4 with the step size 1. Fixing parameter $p=1$, we select the best result of MFSGL and show it in the following part. For compared methods, the value range of parameters are set as they reported and only the best result is given.

\subsubsection{Evaluation Metrics}
For each feature selection method, the samples represent by the selected features are fed into K-means, and the clustering result is used to measure the performance of feature selection. The clustering accuracy(ACC) and the normalized mutual information (NMI) are adopted to evaluate clustering result in this paper \cite{RNMF}.
\par 
ACC reveals the matching degree of clustering result and the ground-truth by discovering the one-to-one correspondence. The definition of ACC is shown as follows:
\begin{equation}\label{ACCeq}
\begin{split}
ACC(h,l)=\frac{\sum_{i=1}^{n}\delta(h_i,l_i)}{n},
\end{split}
\end{equation}
where $h_i$ and $l_i$ are the ground-truth label and clustering result label after best mapping of $i$-th sample respectively. If $h_i=l_i$, $\delta(h_i,l_i)$ equals 1. Otherwise, it equals 0.
\par
Apart from ACC, NMI is another evaluation metric:
\begin{equation}\label{NMIeq}
\begin{split}
NMI=\frac{\sum_{i=1}^{c}\sum_{j=1}^{c}n_{ij} log\frac{n_{ij}}{n_i*\hat{n}_j} }{\sqrt{\sum_{i=1}^{c} n_i log\frac{n_i}{n} * \sum_{j=1}^{c} \hat{n}_j log\frac{\hat{n}_j}{n}}}.
\end{split}
\end{equation}
Here, $n$ is the total number of samples, and $c$ denotes the number of classes. $n_i$ is the number of samples belonging to the $i$-th cluster based on the experimental result, and $\hat{n}_j$ denotes the real number of samples belonging to the $j$-th class. $n_{ij}$ denotes the number of samples which are exist in $i$-th cluster and $j$-th class simultaneously.

\begin{figure}[htbp]
	\centering
	\begin{minipage}[b]{0.48\linewidth}
		\centering
		\centerline{\includegraphics[width=3.9cm]{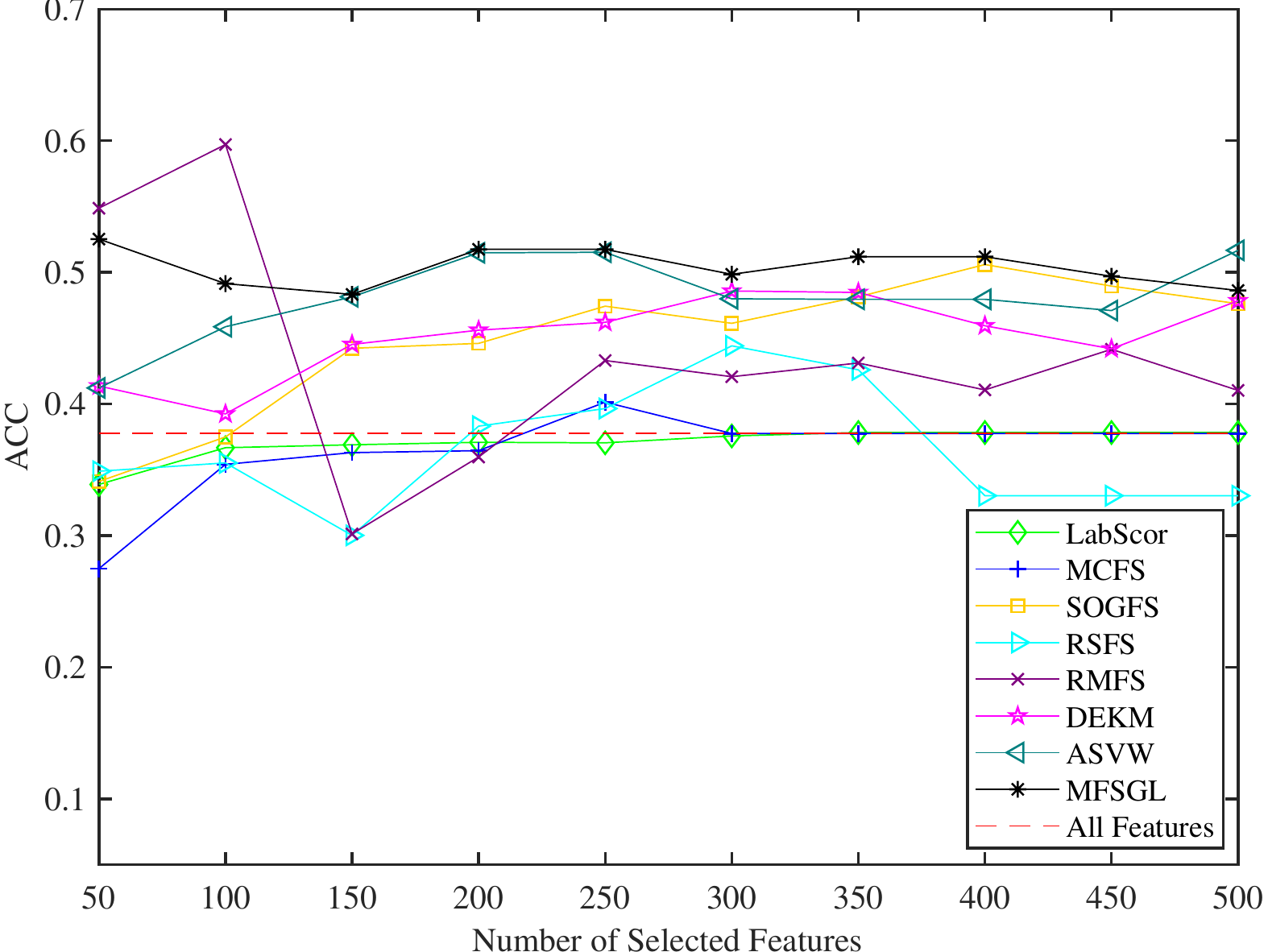}}
		\centerline{Outdoor-Scene}\medskip
	\end{minipage}
	\hfill
	\begin{minipage}[b]{0.48\linewidth}
		\centering
		\centerline{\includegraphics[width=3.9cm]{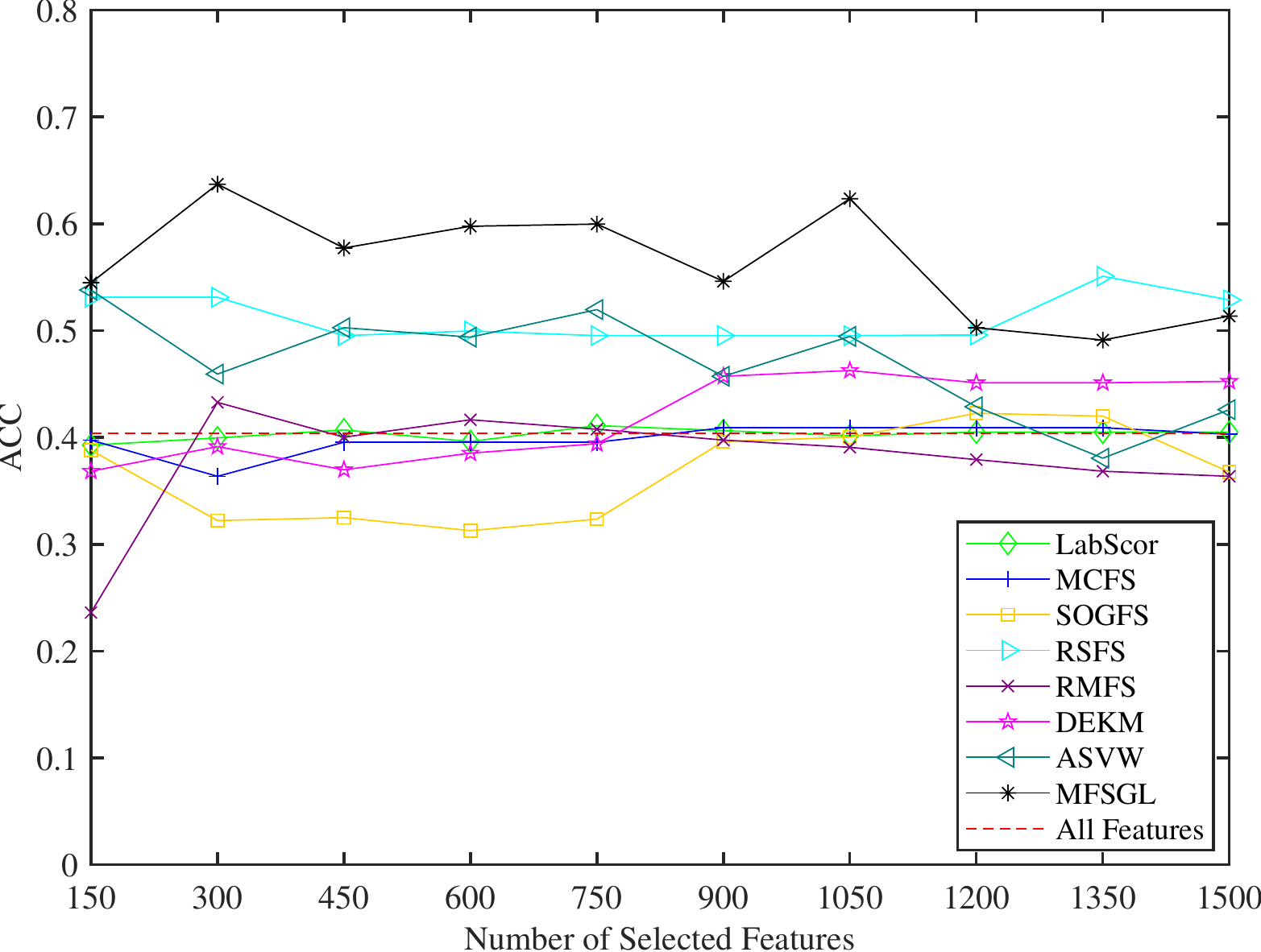}}
		\centerline{Caltech101-7}\medskip
	\end{minipage}
	\begin{minipage}[b]{0.48\linewidth}
		\centering
		\centerline{\includegraphics[width=3.9cm]{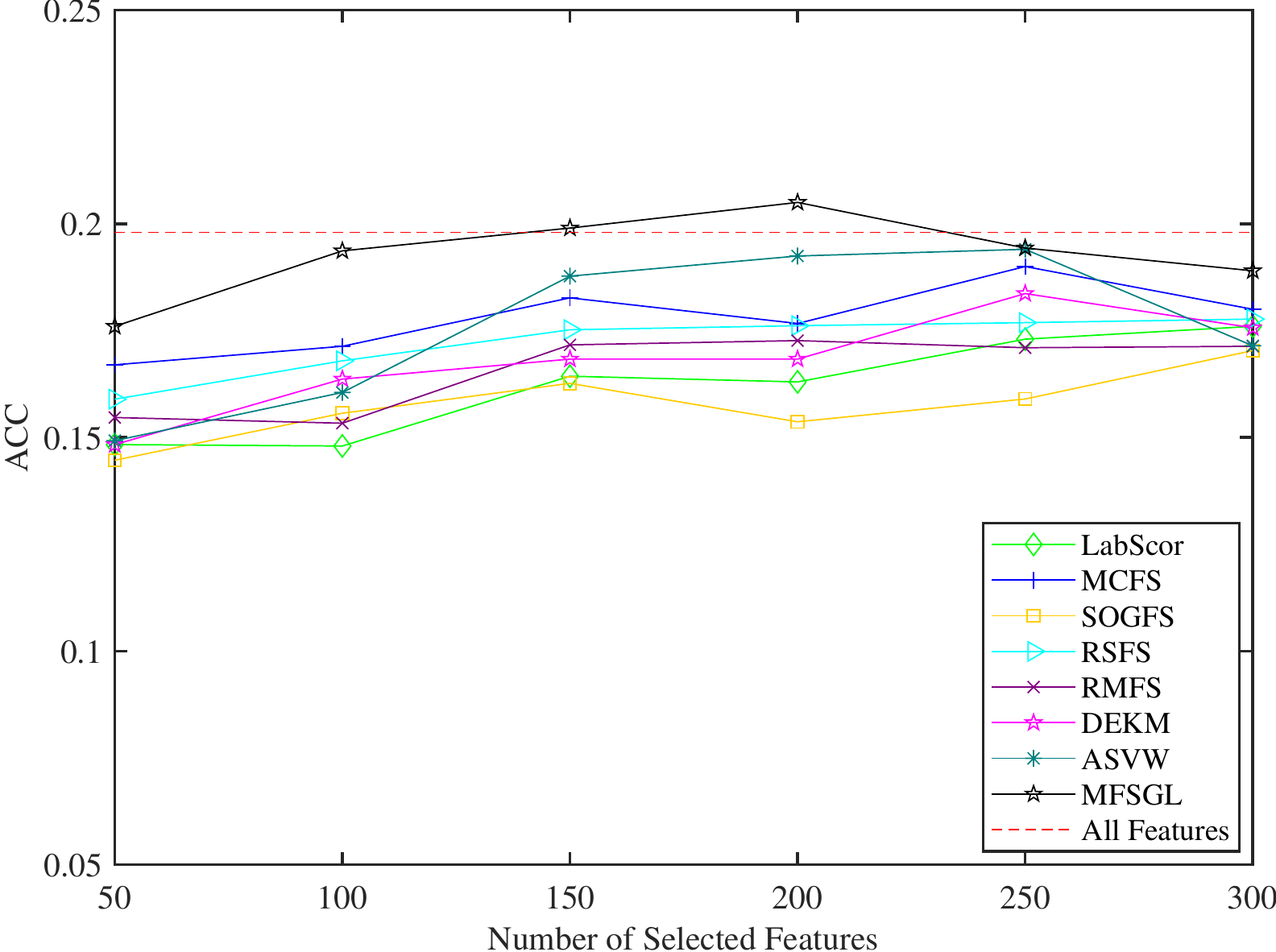}}
		\centerline{NUS-WIDE-OBJ}\medskip
	\end{minipage}
	\hfill
	\begin{minipage}[b]{0.48\linewidth}
		\centering
		\centerline{\includegraphics[width=3.9cm]{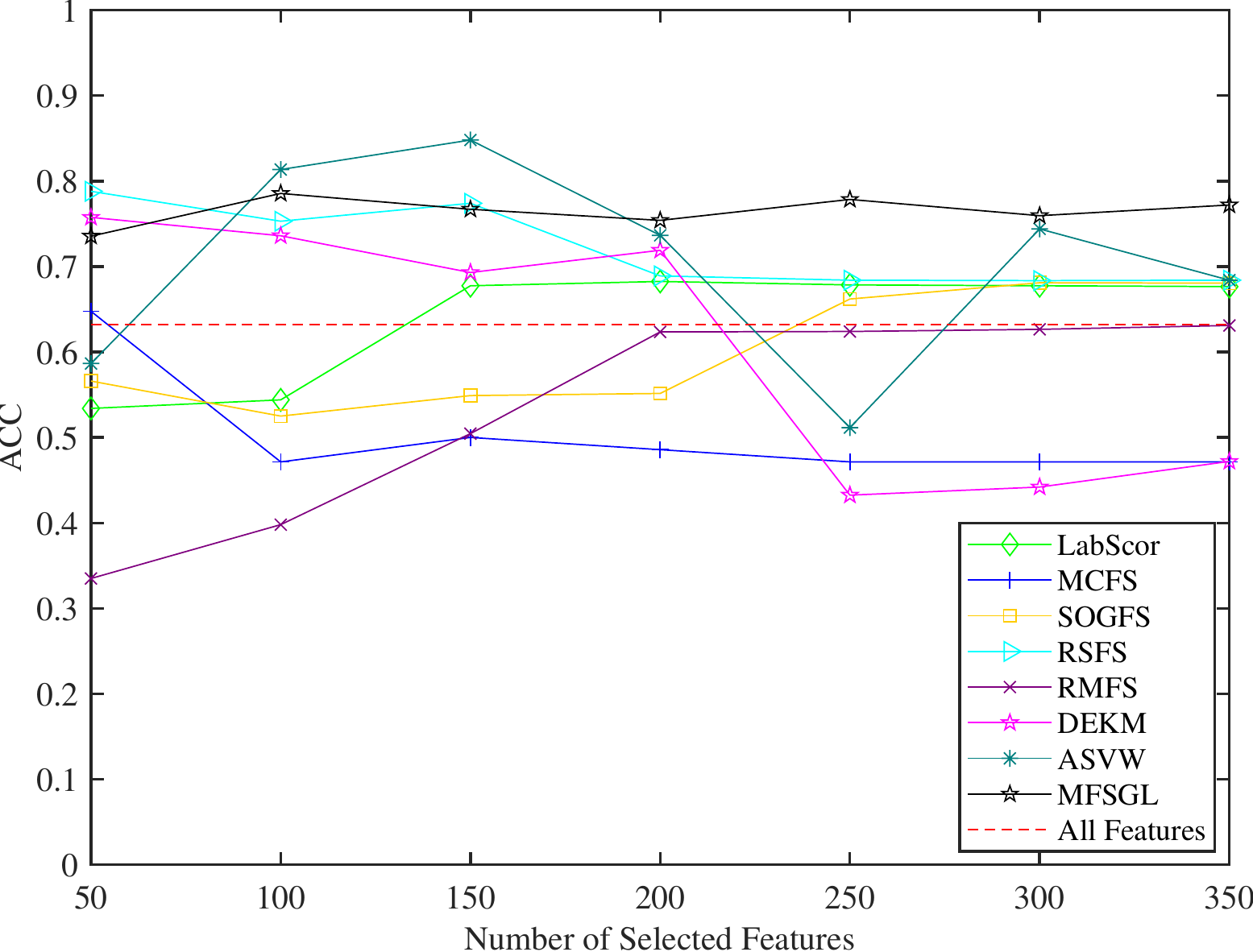}}
		\centerline{Handwritten numerals}\medskip
	\end{minipage}
	\caption{ACC of different methods on four datasets.}
	\label{ACC}
\end{figure}

\begin{figure}[htbp]
	\centering
	\begin{minipage}[b]{0.48\linewidth}
		\centering
		\centerline{\includegraphics[width=3.9cm]{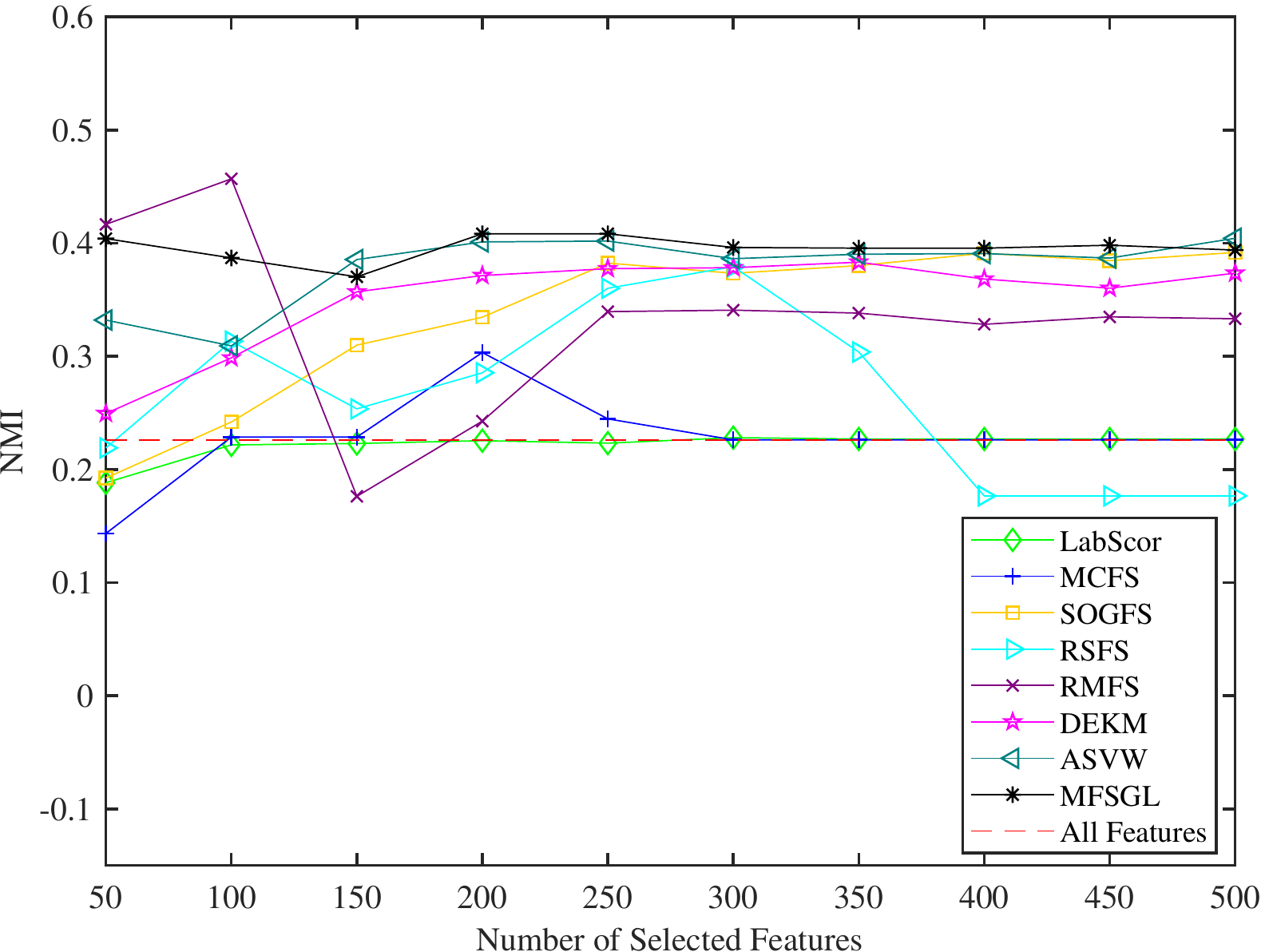}}
		\centerline{Outdoor-Scene}\medskip
	\end{minipage}
	\hfill
	\begin{minipage}[b]{0.48\linewidth}
		\centering
		\centerline{\includegraphics[width=3.9cm]{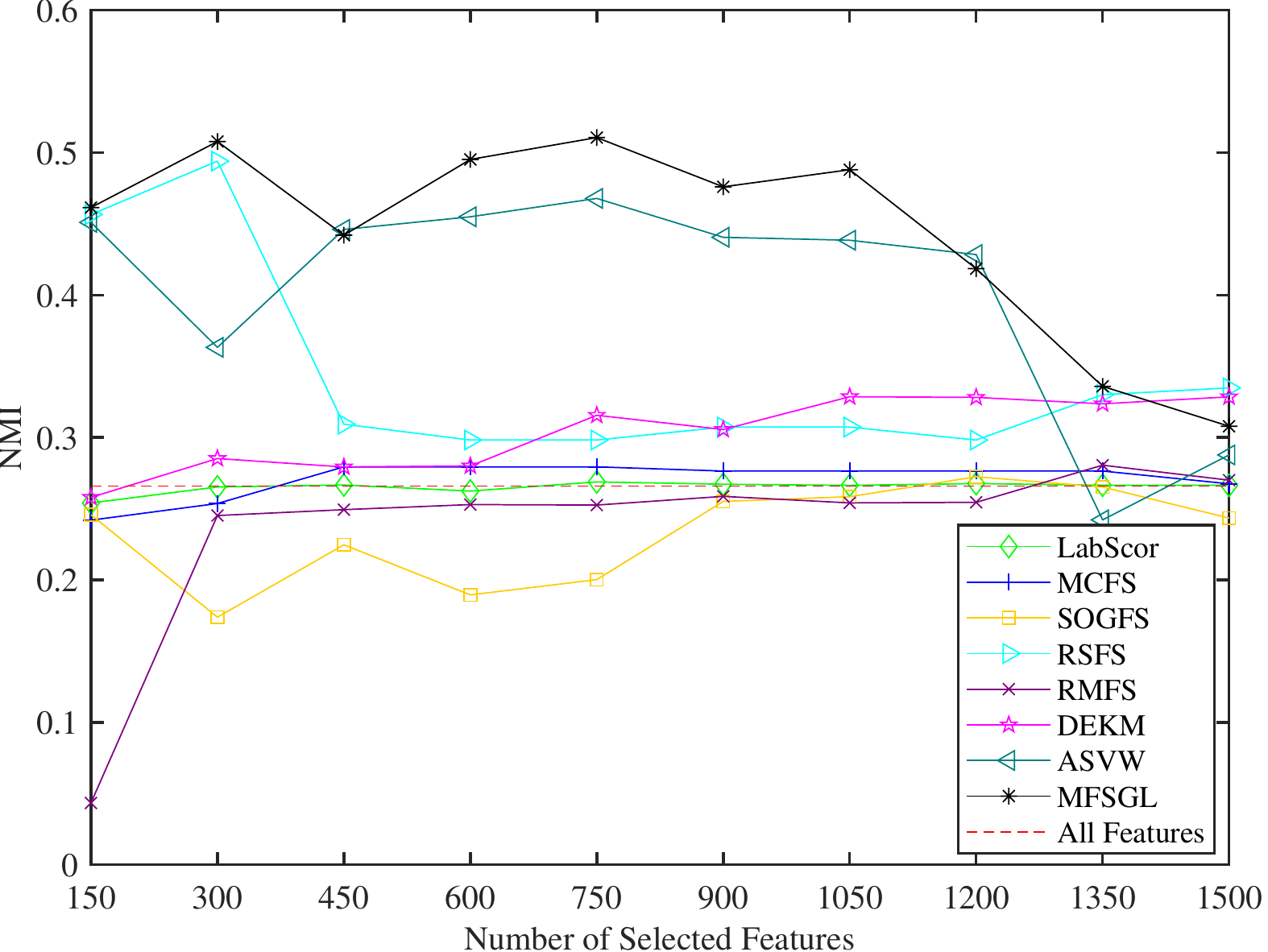}}
		\centerline{Caltech101-7}\medskip
	\end{minipage}
	\begin{minipage}[b]{0.48\linewidth}
		\centering
		\centerline{\includegraphics[width=3.9cm]{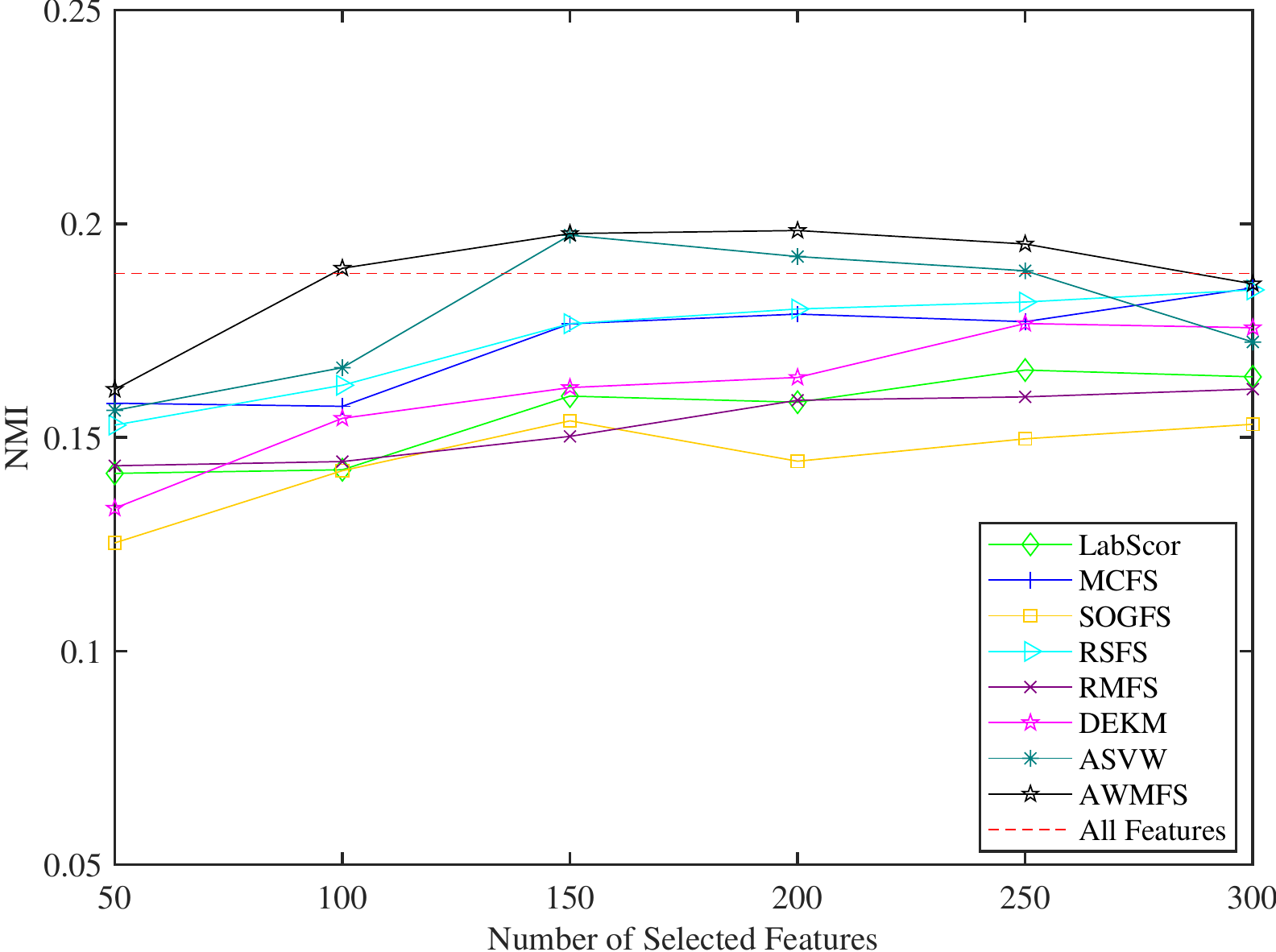}}
		\centerline{NUS-WIDE-OBJ}\medskip
	\end{minipage}
	\hfill
	\begin{minipage}[b]{0.48\linewidth}
		\centering
		\centerline{\includegraphics[width=3.9cm]{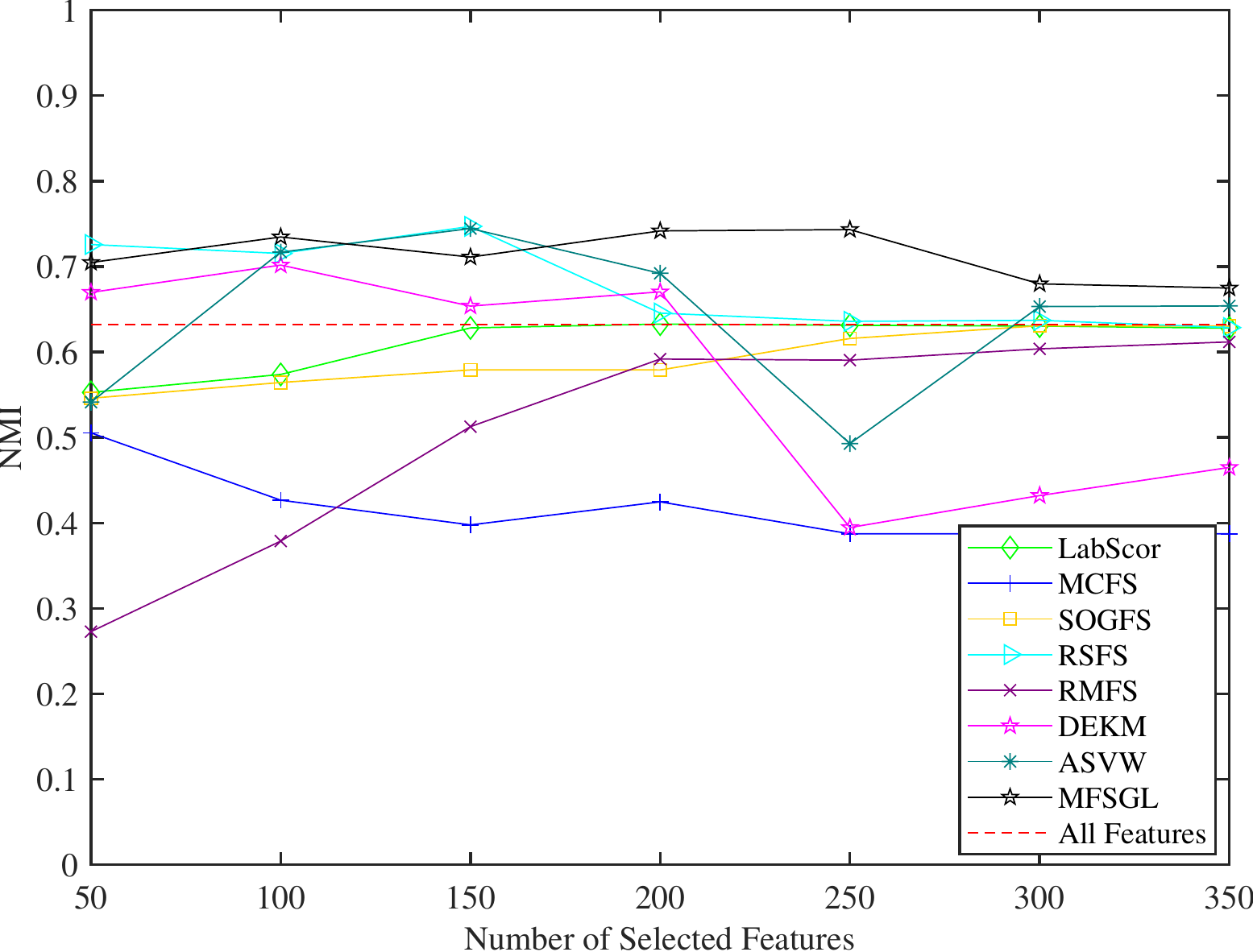}}
		\centerline{Handwritten numerals}\medskip
	\end{minipage}
	\caption{NMI of different methods on four datasets.}
	\label{NMI}
\end{figure}

\begin{figure*}[htb]
	\centering
	\subfigure[ACC with different $\gamma$] {
		\begin{minipage}[b]{0.23\linewidth}
			\centering
			\centerline{\includegraphics[width=4.3cm]{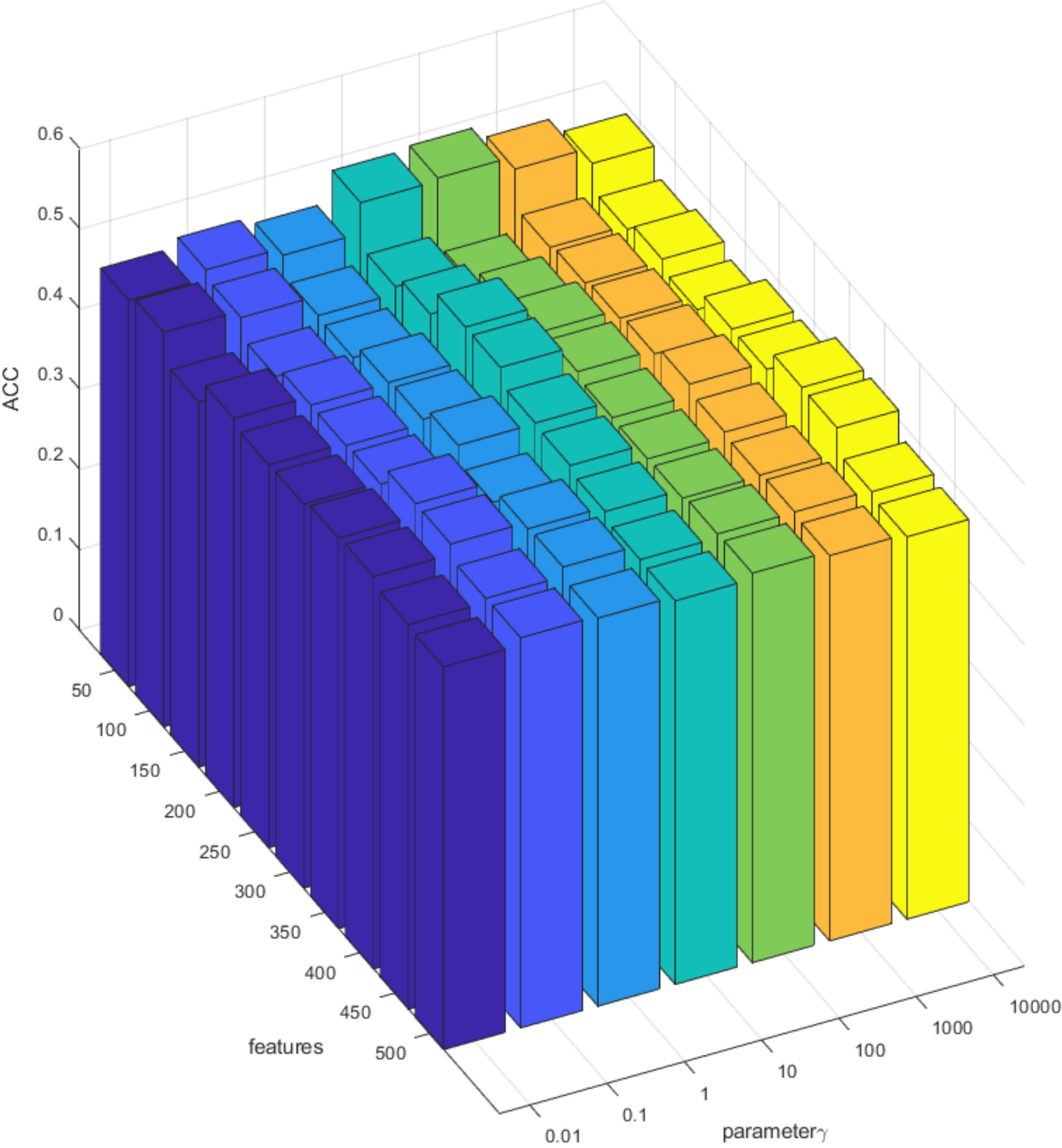}}
			
			\centerline{\scriptsize{Outdoor-Scene}}\medskip
		\end{minipage}
		\hfill	
		\begin{minipage}[b]{0.23\linewidth}
			\centering
			\centerline{\includegraphics[width=4.3cm]{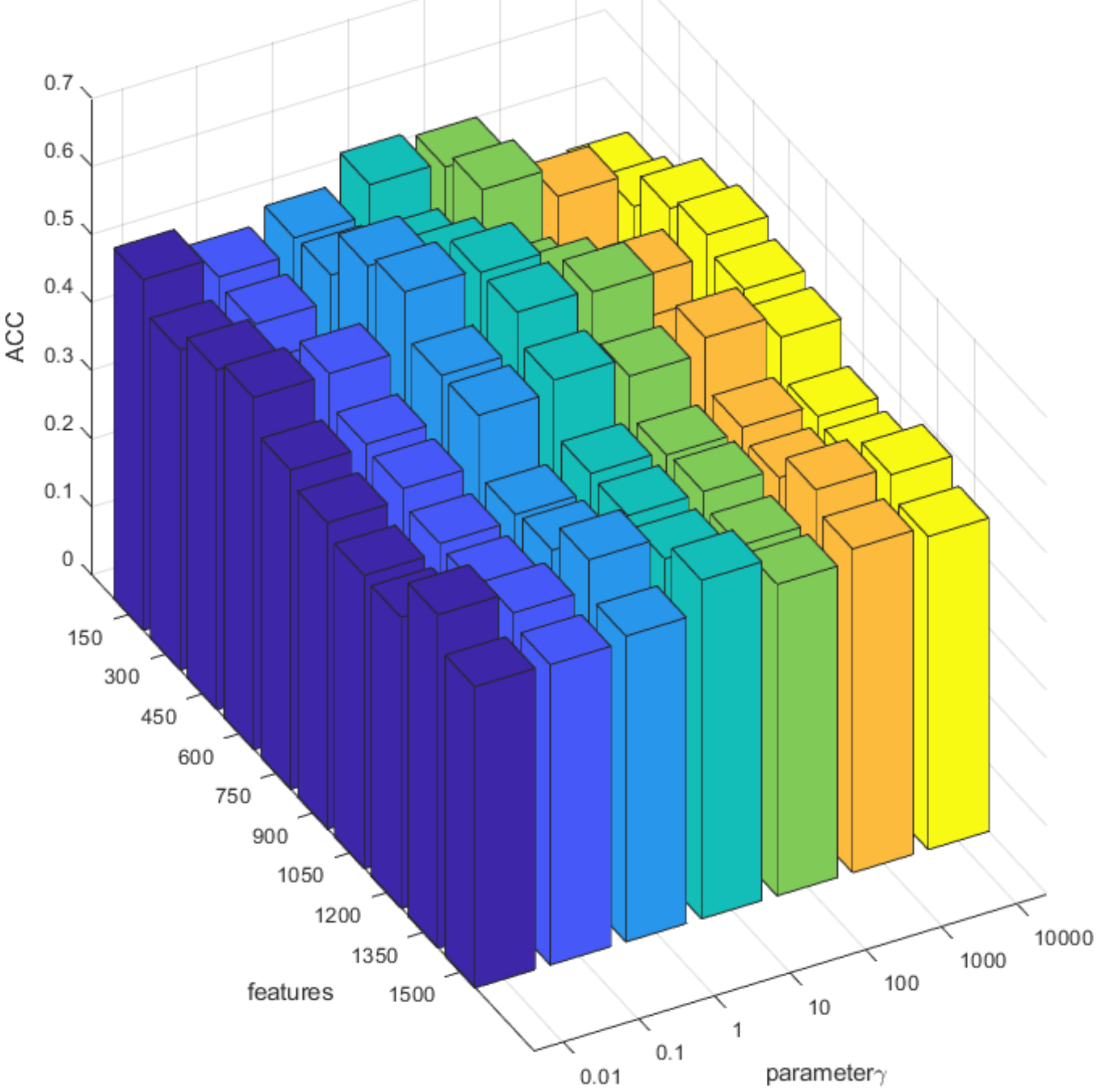}}
			\centerline{\scriptsize{Caltech101-7}}\medskip
		\end{minipage}
		\begin{minipage}[b]{0.23\linewidth}
			\centering
			\centerline{\includegraphics[width=4.3cm]{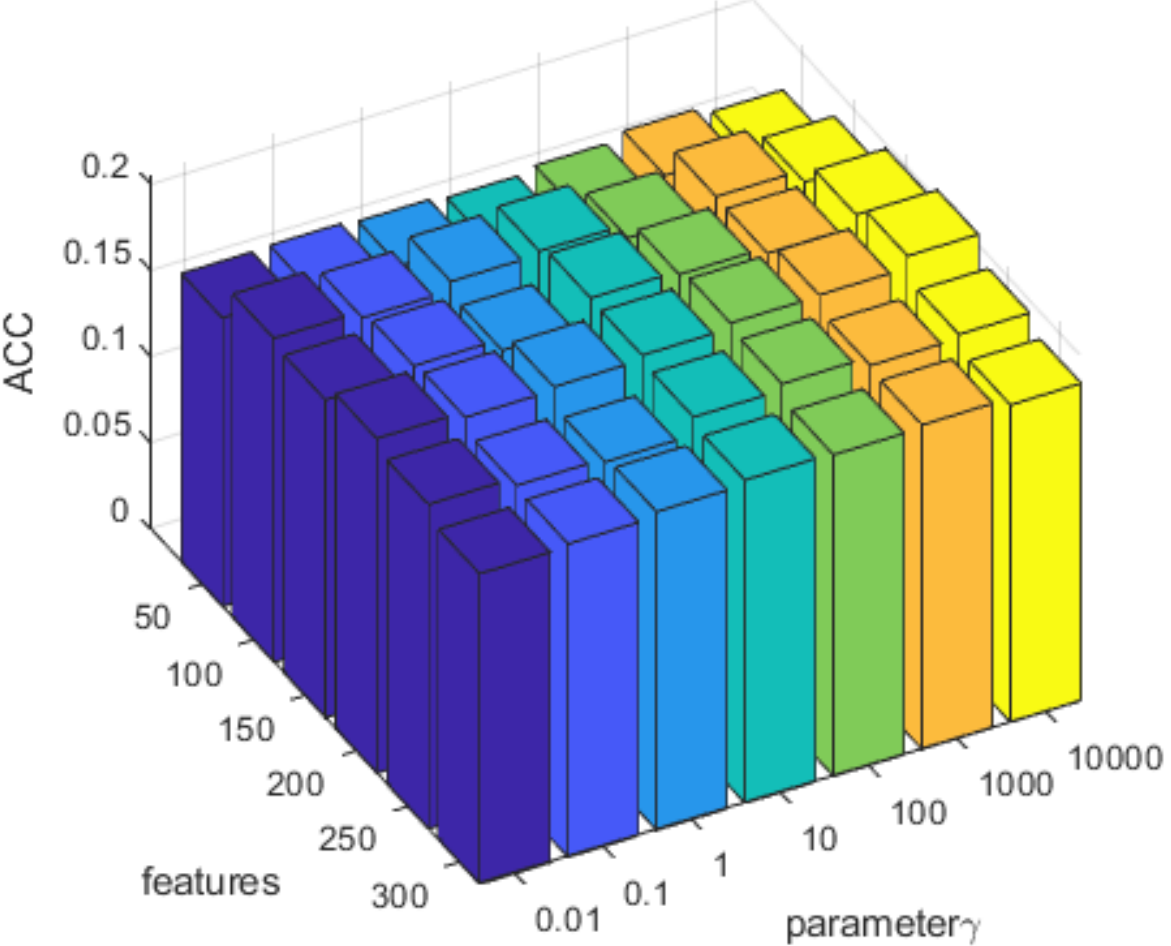}}
			\centerline{\scriptsize{NUS-WIDE-OBJ}}\medskip
		\end{minipage}
		\hfill
		\begin{minipage}[b]{0.23\linewidth}
			\centering
			\centerline{\includegraphics[width=4.3cm]{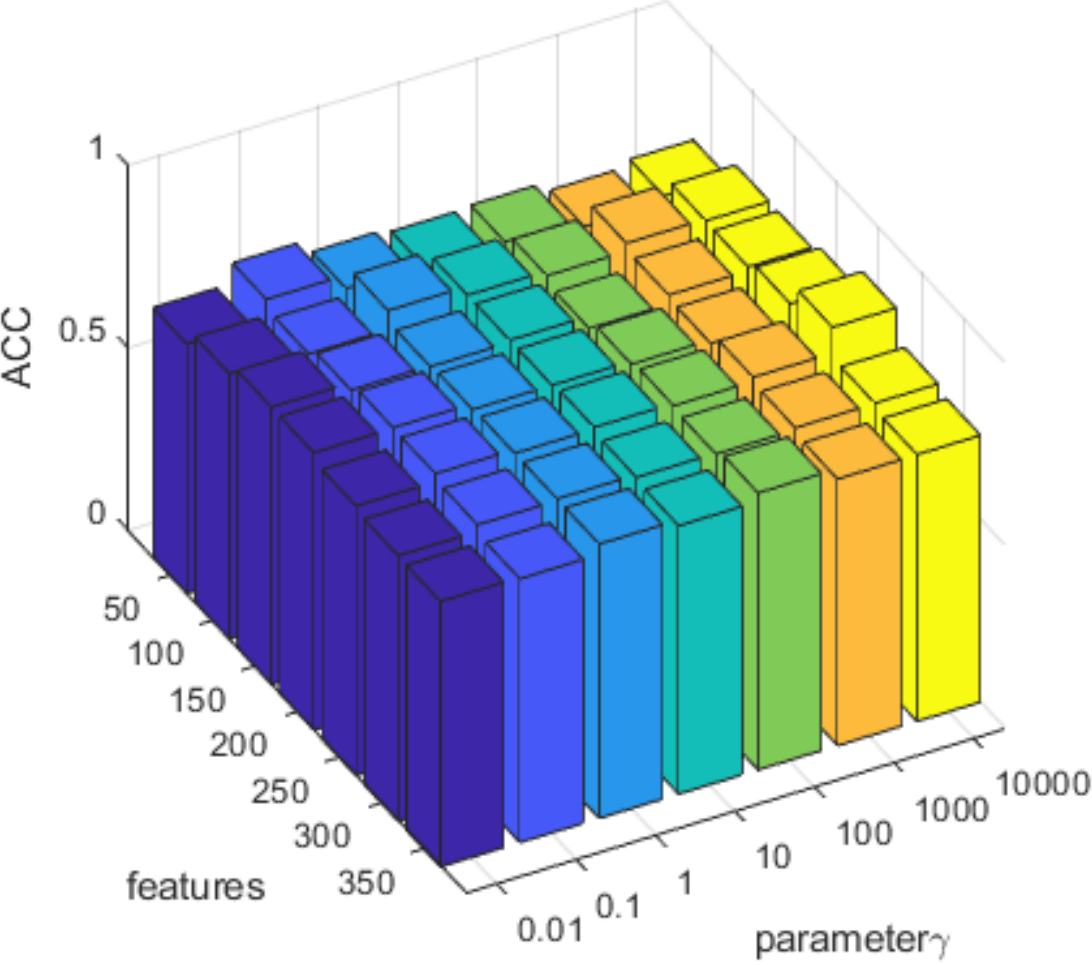}}
			\centerline{\scriptsize{Handwritten numerals}}\medskip
	\end{minipage}}

	\vspace{-0.3cm}
	\subfigure[ACC with different $p$]
	{	
		\begin{minipage}[b]{0.23\linewidth}
			\centering
			\centerline{\includegraphics[width=4.3cm]{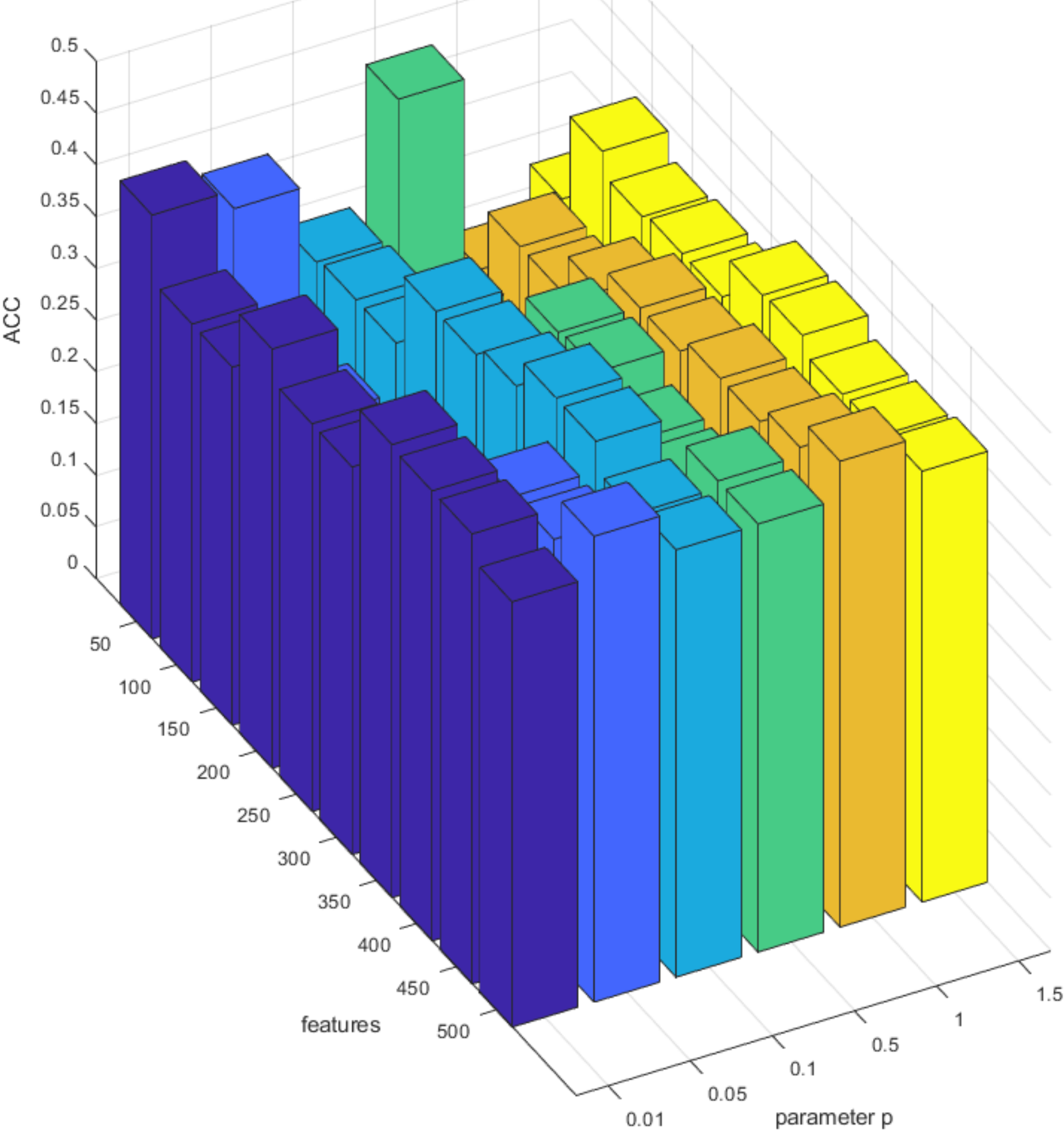}}
			\centerline{\scriptsize{Outdoor-Scene}}\medskip
		\end{minipage}
		\hfill
		\begin{minipage}[b]{0.23\linewidth}
			\centering
			\centerline{\includegraphics[width=4.3cm]{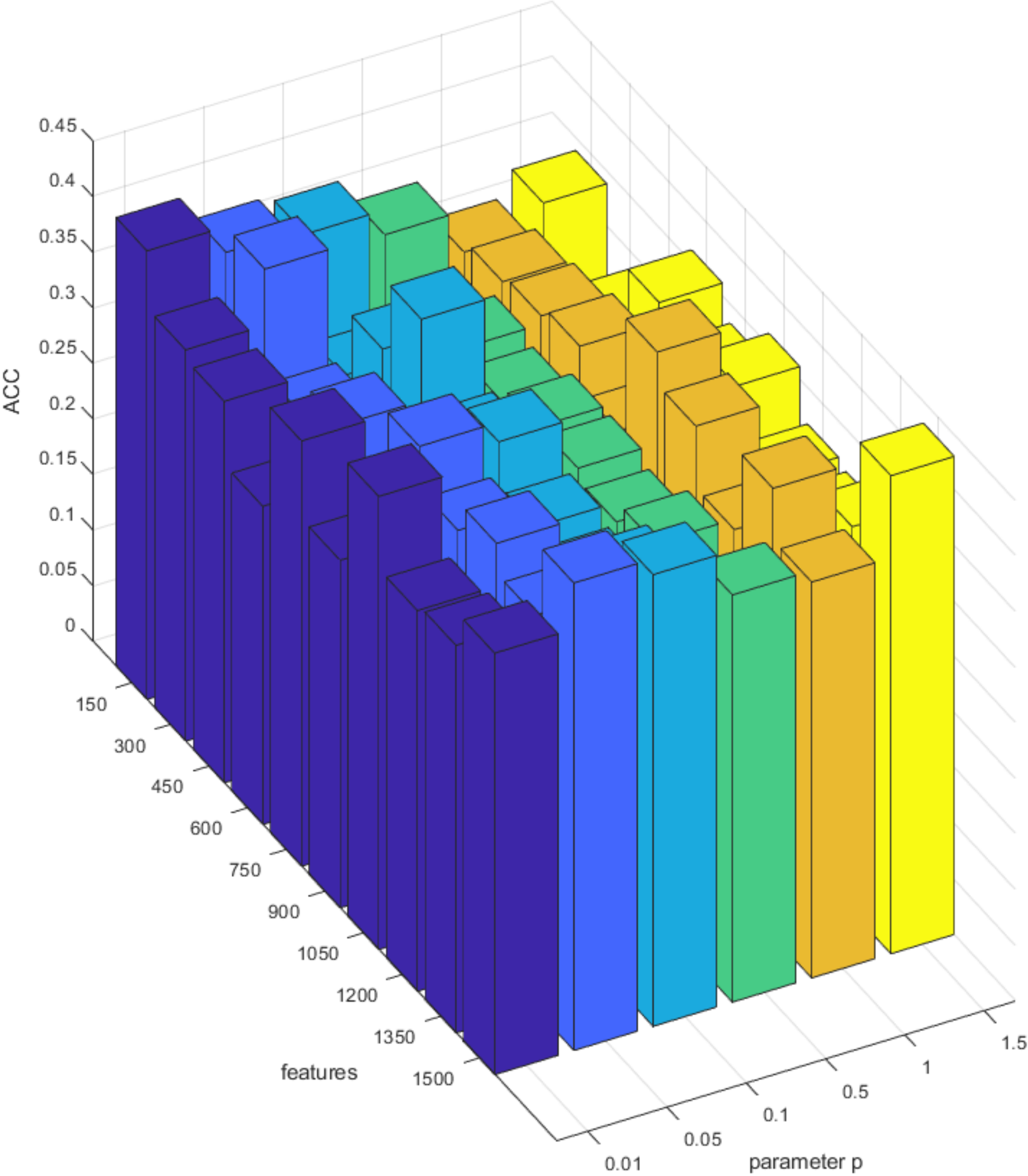}}
			\centerline{\scriptsize{Caltech101-7}}\medskip
		\end{minipage}
		
		\begin{minipage}[b]{0.23\linewidth}
			\centering
			\centerline{\includegraphics[width=4.3cm]{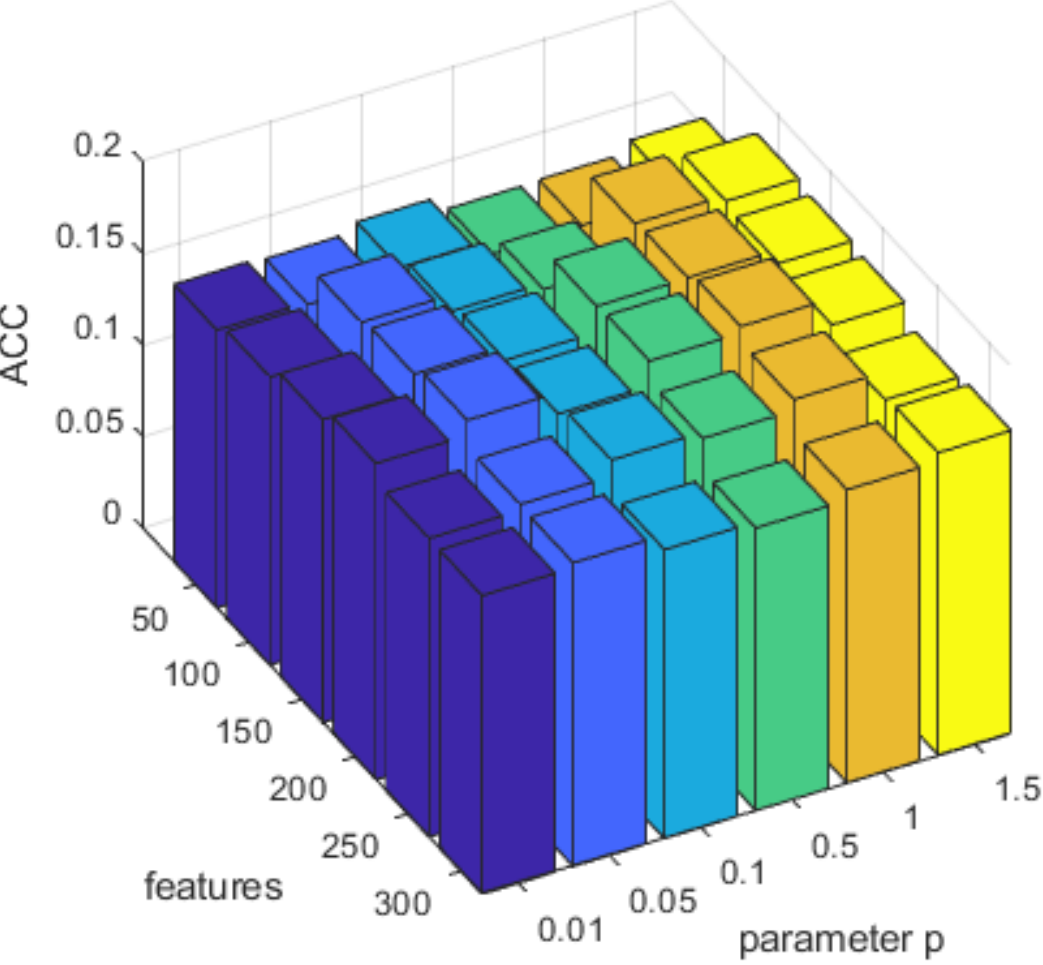}}
			\centerline{\scriptsize{NUS-WIDE-OBJ}}\medskip
		\end{minipage}
		\hfill
		\begin{minipage}[b]{0.23\linewidth}
			\centering
			\centerline{\includegraphics[width=4.3cm]{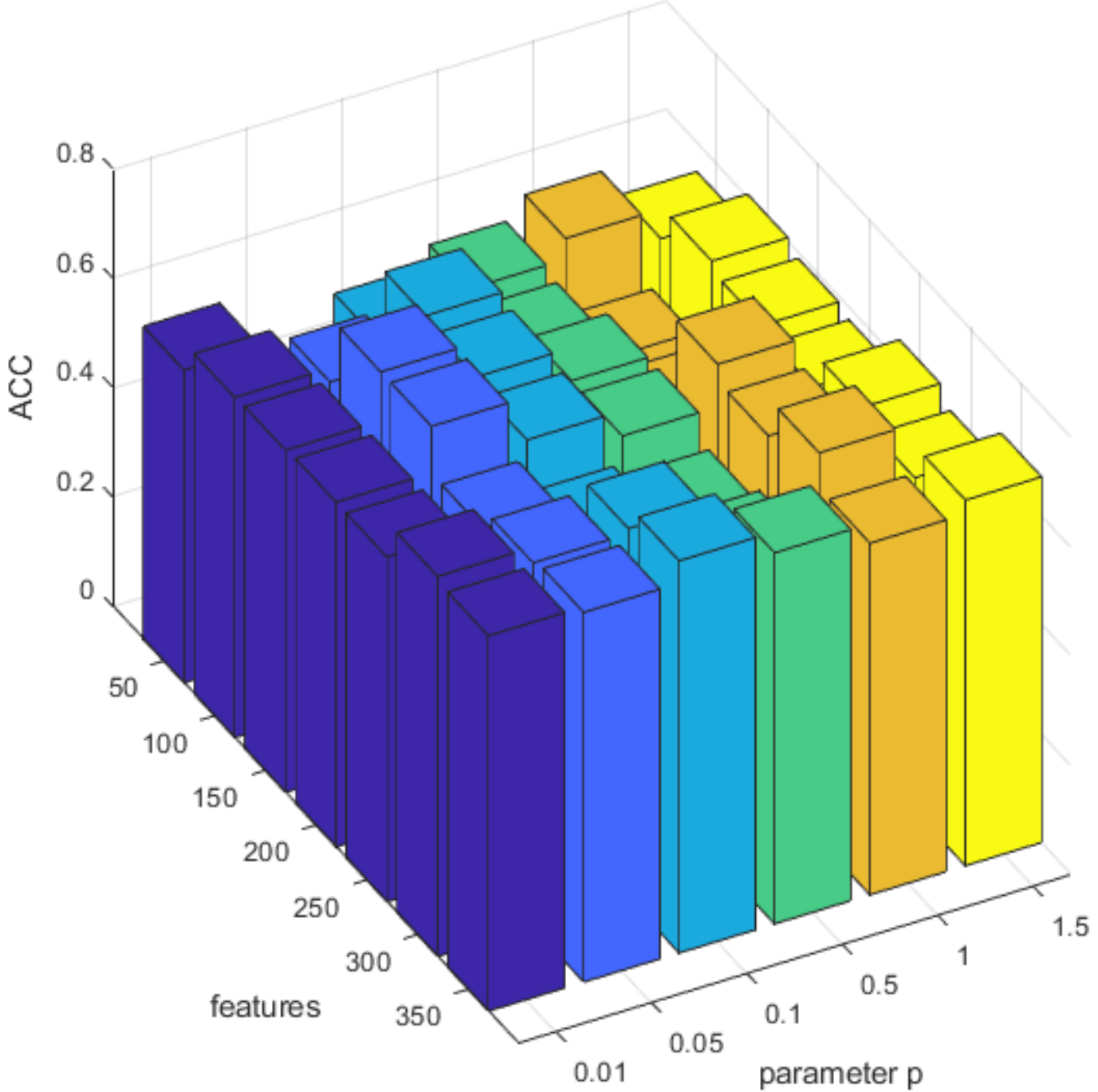}}
			\centerline{\scriptsize{Handwritten numerals}}\medskip
	\end{minipage}}
	
	\caption{Parameter sensitivity study.}
	\label{fig:sensitivity}
\end{figure*}

\subsubsection{Clustering Results with Selected Features}

Fig. \ref{ACC} and Fig. \ref{NMI} show the ACC performance and NMI performance of each approach respectively. The detailed analyses of experimental results are conducted and some valuable points can be obtained.
\begin{itemize}
	\item  As is seen from the experimental results, on \textit{Outdoor-Scene}, \textit{Handwritten numerals} and \textit{Caltech101-7} datasets, the result of MFSGL is superior to other methods in most cases. On \textit{NUS-WIDE-OBJ} dataset, the clustering results of all compared methods are worse than the baseline most of the time. We infer that this is because the feature number of this dataset is too small, and there is not as much noise as others. When the selected subset is too small, it can not contain all the valuable features, and perform worse than the baseline. As the feature number increases, the performance firstly becomes better for the newly added valuable features, and then the performance becomes worse because the feature subset contains too much noise.
	\item In most cases, the performance of multi-view feature selection methods, i.e., MFSGL, ASVW, DEKM and RMFS show more strength than the other single-view feature selection methods. 
	The fact indicates that considering the 
	complementary information across different views makes a big significance in multi-view feature selection. Moreover, we believe that it is the same in other multi-view learning tasks.
	
	\item Obviously, MFSGL and ASVW also outperform the other two multi-view feature selection methods, i.e., RMFS and DEKM. These two methods focus on the global structure. The graph-based methods MFSGL and ASVW try to preserve the underlying local manifold structure shared by all views. It reveals that the local structure information is superior to the global structure information once again, which has been widely recognized by researchers.
	\item  Furthermore, the proposed MFSGL also outperforms ASVW. Both MFSGL and ASVW try to learn the common similarity matrix of all views. Different from ASVW, the similarity matrix learned by MFSGL has an ideal neighbor assignment owing to the added rank constraint, which contributes a lot to the superior performance.
\end{itemize}

\begin{figure}[htb] \label{conv}
	\centering
	\begin{minipage}[b]{0.43\linewidth}
		\centering
		\centerline{\includegraphics[width=4.0cm]{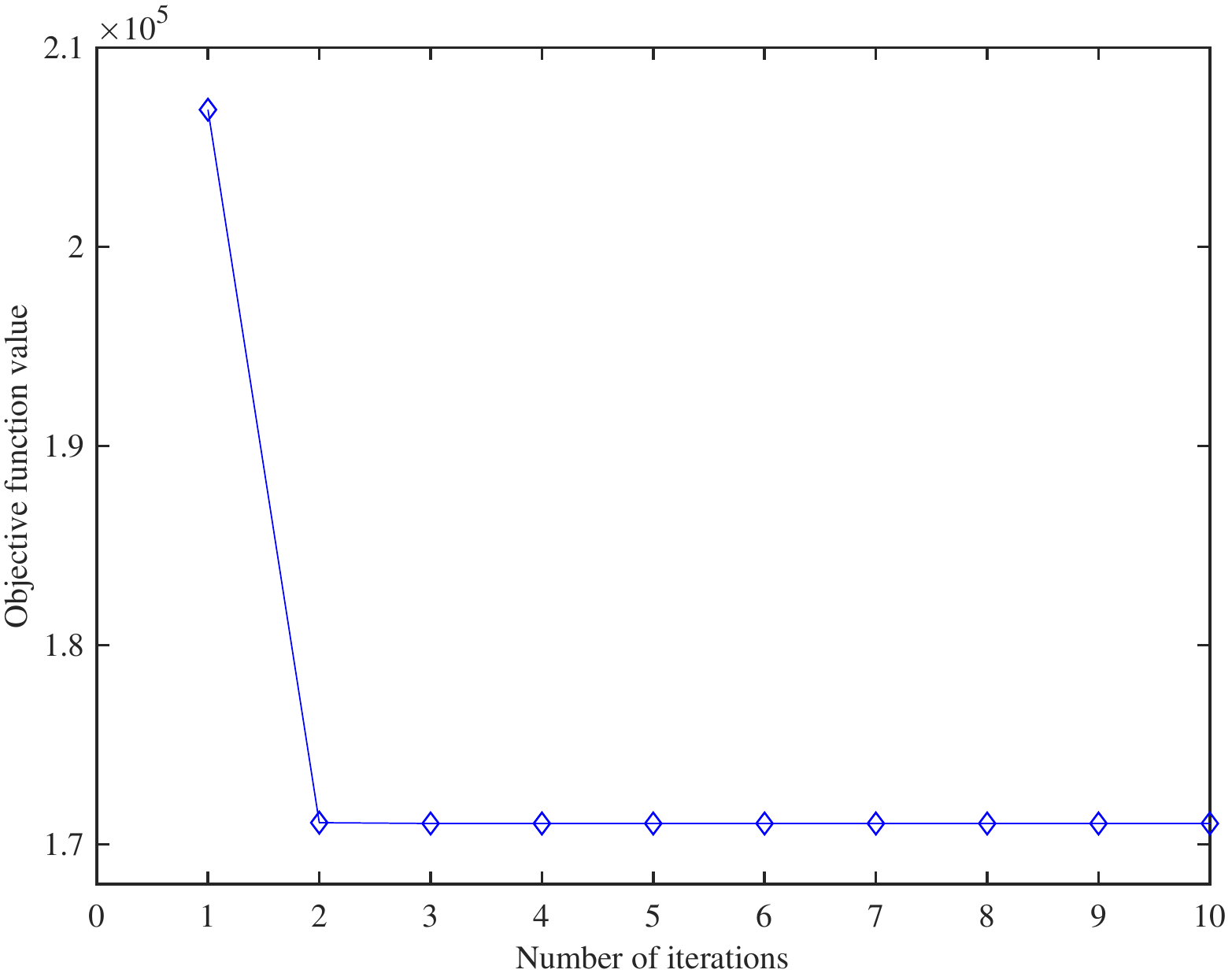}}
		\centerline{Outdoor-Scene}\medskip
	\end{minipage}
	\hfill
	\begin{minipage}[b]{0.43\linewidth}
		\centering
		\centerline{\includegraphics[width=4.0cm]{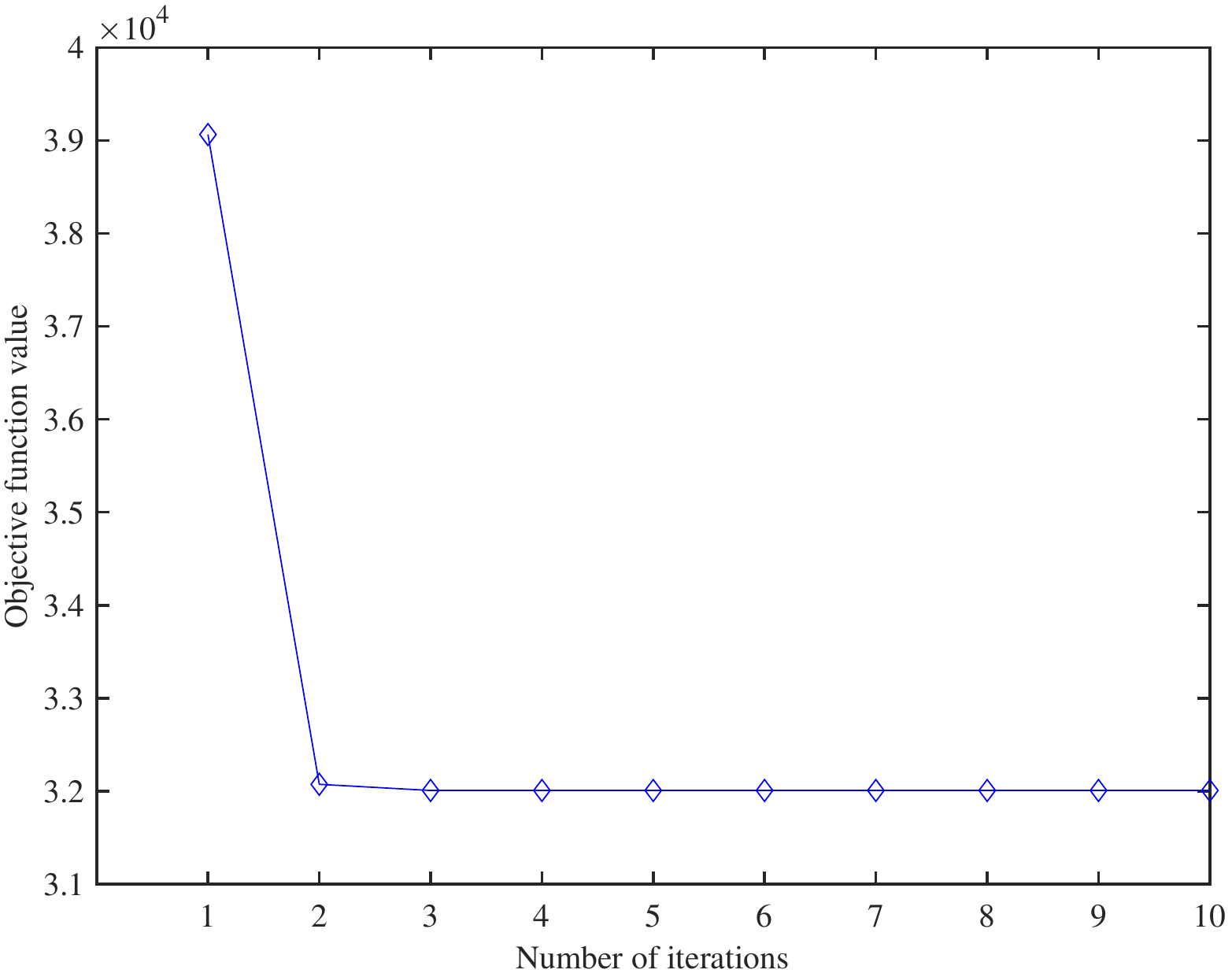}}
		\centerline{Caltech101-7}\medskip
	\end{minipage}
	\begin{minipage}[b]{0.43\linewidth}
		\centering
		\centerline{\includegraphics[width=4.0cm]{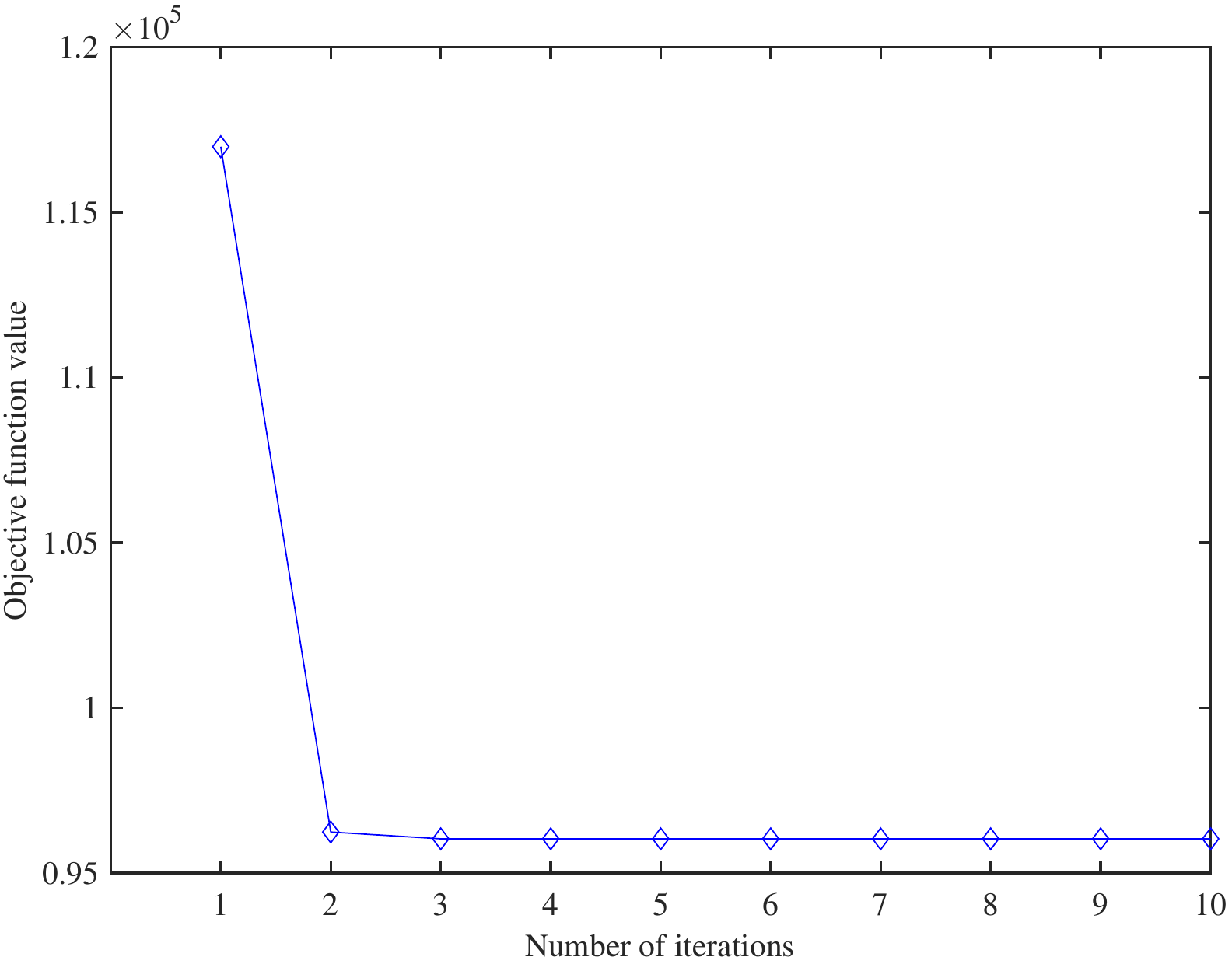}}
		\centerline{NUS-WIDE-OBJ}\medskip
	\end{minipage}
	\hfill
	\begin{minipage}[b]{0.43\linewidth}
		\centering
		\centerline{\includegraphics[width=4.0cm]{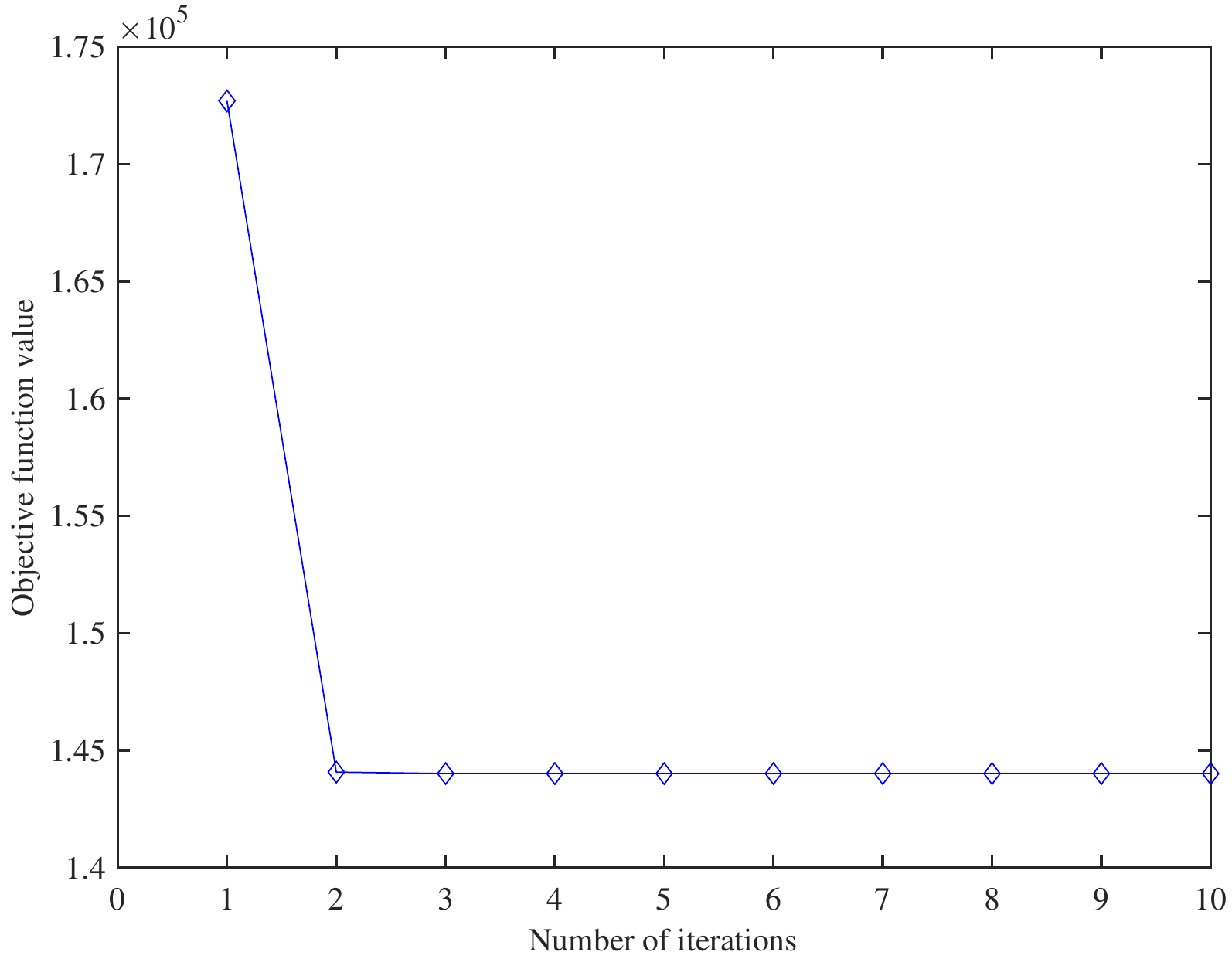}}
		\centerline{Handwritten numerals}\medskip
	\end{minipage}

	\caption{Convergence curves of Algorithm 1 on four datasets.}
	\label{objw}
\end{figure}

%

\subsubsection{Parameter Sensitivity Study}
The parameter $ m_v$ gives slight influence on the performance, and in our experiments we set it around $\frac{d_v}{3}$ to $\frac{2d_v}{3}$. This is because if $m_v$ is set too small, some valuable features may be lost, and if $m_v$ is set too large, the learned feature subset may be not compact and discriminative enough. On account of the parameter $\gamma$ controls the row sparsity of  $ W_v$, we concentrate on the performance of MFSGL with various $\gamma$. With fixed $p=1$, the best ACC results under various $\gamma$ on four real-world datasets are shown in Fig. 7a. It is easy to know that the proposed MFSGL is robust to $\gamma$ to a certain extent. In spite of this, it is suggested to search the optimal $\gamma$ for the best performance in practical application.

\par 
Moreover, the influence of different exponential functions should also be taken into account. Therefore, we change the value of parameter $p$ and fix other parameters, and the variance on performance of MFSGL is demonstrated in Fig. 7b. It is clear that there is no significant difference with different $p$, and we can easily conclude that the performances of MFSGL with different exponential functions are all at a high level.

\subsubsection{Convergence Study}
The Algorithm 1 is proposed to solve problem (\ref{W2}) and its convergence has been proved in Section \ref{subsec: Convergence Analysis of Alogrithm 1}. Here, the speed of its convergence is further studied by experiment. For brevity, we only show the result of one view on each dataset. In Fig. \ref{objw}, it is obvious that Algorithm 1 converges within about 5 iterations. And it also keeps the same convergence speed in other views. The fast convergence saves much computational time of the proposed feature selection framework.

\section{Conclusion}
\label{sec:Conclusion}
In this paper, we present a novel method called MFSGL for unsupervised multi-view feature selection. MFSGL learns an optimal similarity graph across different views by adding a reasonable constraint, and an efficient view weight assignment strategy is adopted to balance the contribution of each view. An algorithm is also given to optimize this problem. The experiments on the synthetic datasets and four benchmark real-world datasets validate the superiority of MFSGL. In future, we will be committed to find some new techniques to further improve the quality of similarity graph.


%


\ifCLASSOPTIONcaptionsoff
  \newpage
\fi

\bibliographystyle{IEEEtran}
\bibliography{ref}



%

%

\begin{IEEEbiography}[{\includegraphics[width=1in,height=1.25in,clip,keepaspectratio]{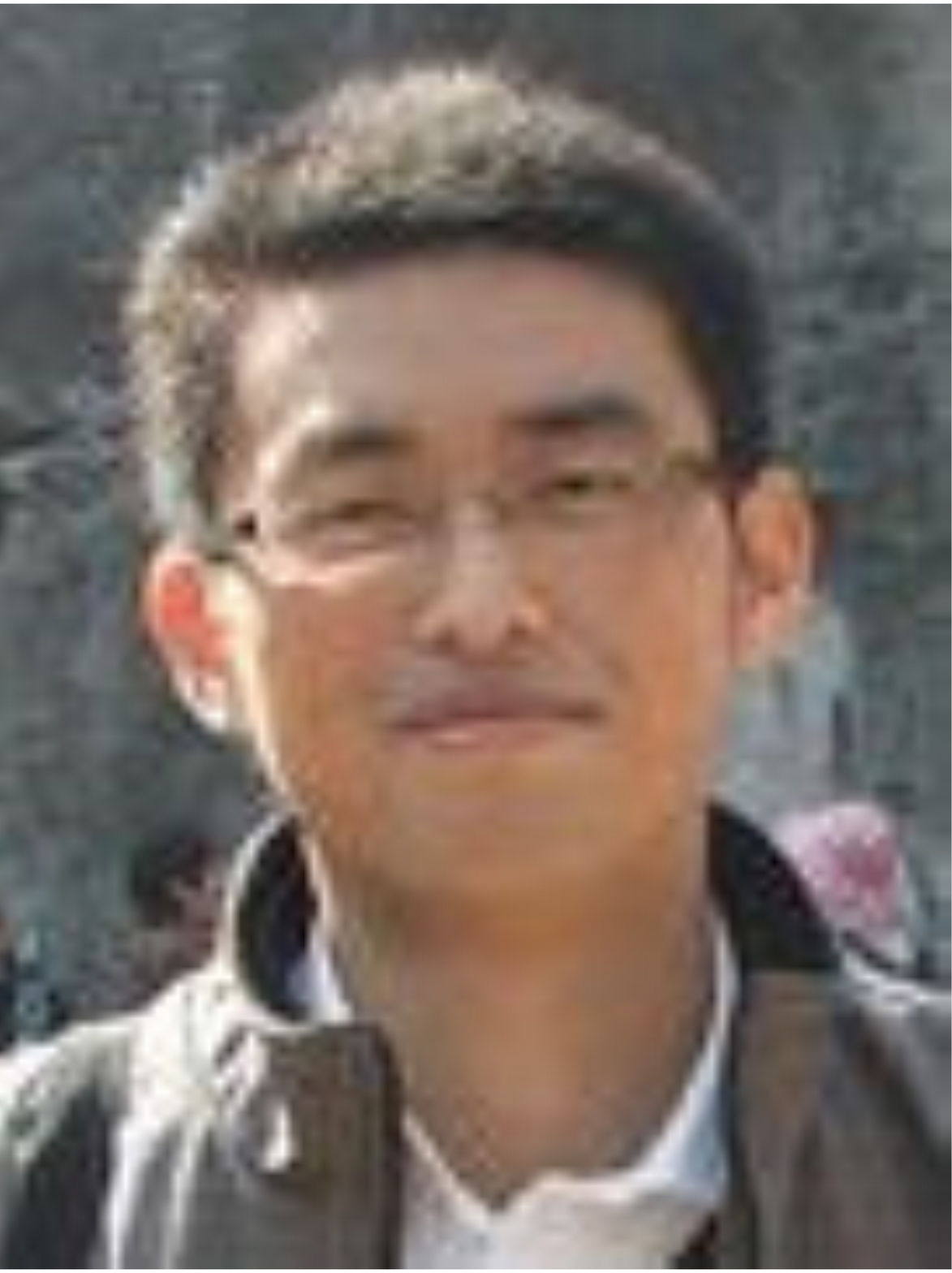}}]{Qi Wang}
	(M'15-SM'15) received the B.E. degree in automation and the Ph.D. degree in pattern recognition and intelligent systems from the University of Science and Technology of China, Hefei, China, in 2005  and 2010, respectively.  He is currently a Professor with the School of Artificial Intelligence, Optics and Electronics (iOPEN), Northwestern Polytechnical University, Xi'an, China. His research interests include computer vision, pattern recognition and remote sensing.
\end{IEEEbiography}

\begin{IEEEbiography}[{\includegraphics[width=1in,height=1.25in,clip,keepaspectratio]{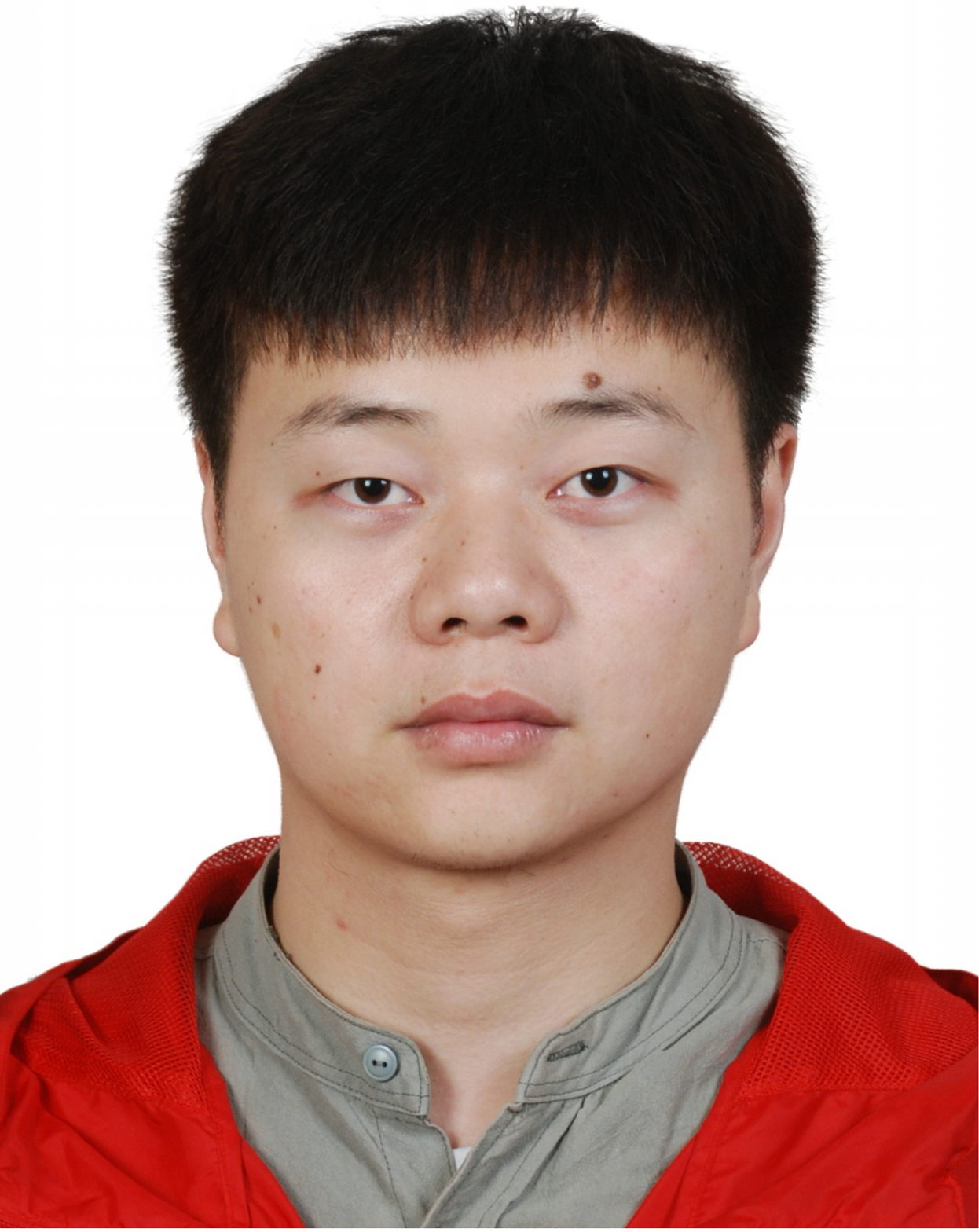}}]{Xu Jiang}
	received the B.E. degree in electronics and information engineering from Northwestern Polytechnical University, Xi'an, China, in 2019. He is currently working toward the M.S. degree in computer science in the School of Artificial Intelligence, Optics and Electronics (iOPEN), Northwestern Polytechnical University, Xi'an, China. His research interests include machine learning and data mining.
\end{IEEEbiography}

\begin{IEEEbiography}[{\includegraphics[width=1in,height=1.25in,clip,keepaspectratio]{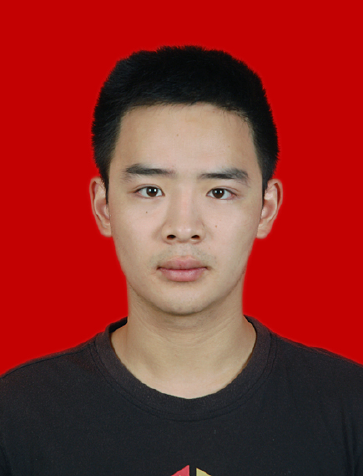}}]{Mulin Chen}
	received the B.E. degree in software engineering and the Ph.D. degree in computer application technology from Northwestern Polytechnical University, Xi’an, China, in 2014 and 2019 respectively. He is currently a researcher with the School of Artificial Intelligence, Optics and Electronics (iOPEN), Northwestern Polytechnical University, Xi'an, China. His current research interests include computer vision and machine learning.
\end{IEEEbiography}

\begin{IEEEbiographynophoto}{Xuelong Li}
	(M'02-SM'07-F'12) is a full professor with the School of Artificial Intelligence, Optics and Electronics (iOPEN), Northwestern Polytechnical University, Xi'an, China.
\end{IEEEbiographynophoto}




\end{document}